%% file: main.tex
\documentclass[sigconf]{acmart}
\AtBeginDocument{%
  }

\setcopyright{acmlicensed} 
\copyrightyear{2026} 
\acmYear{2026} 
\acmDOI{XXXXXXX.XXXXXXX} 
\acmConference[Conference acronym 'XX]{}{June 03--05,
  2018}{Woodstock, NY}  
\acmISBN{978-1-4503-XXXX-X/2018/06}  

\input{preamble}
\newcommand{\sysname}{\textsc{Mirage}\xspace}

\begin{document}

\title{Systematic Discovery of Semantic Attacks in Online Map Construction through Conditional Diffusion}


\author{Chenyi Wang}
\affiliation{%
  \institution{University of Arizona}
  \country{}}
\email{chenyiw@arizona.edu}

\author{Ruoyu Song}
\affiliation{%
  \institution{Purdue University}
  \country{}}
\email{song464@purdue.edu}

\author{Raymond Muller}
\affiliation{%
  \institution{Lawrence Livermore National Laboratory}
  \country{}}
\email{muller15@llnl.gov}

\author{Jean-Philippe Monteuuis}
\affiliation{%
  \institution{Qualcomm}
  \country{}}
\email{jmonteuu@qti.qualcomm.com}

\author{Jonathan Petit}
\affiliation{%
  \institution{Qualcomm}
  \country{}}
\email{petit@qti.qualcomm.com}

\author{Z. Berkay Celik}
\affiliation{%
  \institution{Purdue University}
  \country{}}
\email{zcelik@purdue.edu}

\author{Ryan Gerdes}
\affiliation{%
  \institution{Virginia Tech}
  \country{}}
\email{rgerdes@vt.edu}

\author{Ming F. Li}
\affiliation{%
  \institution{University of Arizona}
  \country{}}
\email{lim@arizona.edu}
\renewcommand{\shortauthors}{C. Wang, R. Song, R. Muller, J.P. Monteuuis, J. Petit, Z.B. Celik, R. Gerdes, M.F. Li}



\input{secs/abs}

\maketitle

\input{secs/intro}

\input{secs/related_work}

\input{secs/threat_model}

\input{secs/methodology}

\input{secs/evaluation}

\input{secs/case_study}
\input{secs/discussion}
\input{secs/conclusion}

\begin{acks}
This work used NCSA Delta GPU at UIUC through allocation CIS260379 from the Advanced Cyberinfrastructure Coordination Ecosystem: Services \& Support (ACCESS) program, which is supported by U.S. National Science Foundation grants \#2138259, \#2138286, \#2138307, \#2137603, and \#2138296. We thank University of Arizona Tech Park and Dhia Neifar for helping with the physical experiments.
\end{acks}

\bibliographystyle{ACM-Reference-Format}
\bibliography{references}


\appendix

\section{Generative AI Usage}

During the preparation of this paper, the authors used Google Gemini and Anthropic Claude for language refinement (grammar, clarity, concision, and organization), and Anthropic Claude Code for implementation assistance (code completion, debugging, and refactoring). 
All AI-assisted edits and code modifications were subsequently reviewed and verified by the authors, who take full responsibility for the accuracy, originality, and correctness of the work.

\section{Open Science}
\label{app:openscience}

We release the following artifacts in the following anonymized repo for public use and validation \url{https://github.com/WiSeR-Lab/MIRAGE}, which include the following:
\begin{itemize}
\item \textbf{Attack code}: Full implementation of \sysname (latent inversion, PGD optimization, CLIP direction loss, boundary injection), pixel PGD baseline, and AdvPatch baseline reimplementation.
\item \textbf{Evaluation code}: Defense evaluation pipeline (all 6 families), A* planner with ORR/FSR computation.
\item \textbf{Model checkpoints}: Links to publicly available MapTR and MagicDrive weights.
\end{itemize}
The nuScenes dataset is publicly available. 

\input{secs/ethics}


\input{secs/appendix}

\end{document}

%% file: preamble.tex
\usepackage{amsmath}
\usepackage{booktabs}
\usepackage{subcaption}
\usepackage{multirow}
\usepackage{algorithm}
\usepackage{algorithmic}
\usepackage{xspace}
\usepackage{graphicx}
\usepackage{tikz}
\usepackage{pgfplots}
\usepackage[table]{xcolor}
\pgfplotsset{compat=1.18}
\usepgfplotslibrary{statistics}

\def\eg{{e.g.},~}

\DeclareRobustCommand*\circled[1]{\tikz[baseline=(char.base)]{ \node[shape=circle,draw,color=white,fill=black,inner sep=0.5pt] (char){#1};}}

%% file: secs/abs.tex
\begin{abstract}
Autonomous vehicles depend on online HD map construction 
to perceive lane boundaries, dividers, and pedestrian crossings---safety-critical road elements that directly govern motion planning. While existing pixel perturbation attacks can disrupt the mapping, they can be neutralized by standard adversarial defenses. We present \sysname, a framework for systematic discovery of semantic attacks that bypass adversarial defenses and degrade mapping predictions by finding plausible environmental variation (\eg shadows, wet roads). \sysname exploits the latent manifold of real-world data learned by diffusion models, and searches for semantically mutated scenes neighboring the ground truth with the same road topology yet mislead the mapping predictions. 
%
%
We evaluate \sysname on nuScenes and demonstrate two attacks: (1)~\emph{boundary removal}, suppressing 57.7\% of detections and corrupting 96\% of planned trajectories; and (2)~\emph{boundary injection}, the only method that successfully injects fictitious boundaries, while pixel PGD and AdvPatch fail entirely. Both attacks remain potent under various adversarial defenses. 
We use two independent VLM judges to quantify realism, where
\sysname passes as realistic 80--84\% of the time (vs.\ 97--99\% for clean nuScenes), while 
AdvPatch only 0--9\%. 
Our findings expose a categorical gap in current adversarial defenses: semantic-level perturbations that manifest as legitimate environmental variation are substantially harder to mitigate than pixel-level perturbations.
\end{abstract}

%% file: secs/intro.tex
\section{Introduction}
\label{sec:intro}

\begin{figure}[ht]
  \centering
  \begin{subfigure}[t]{0.33\columnwidth}
      \includegraphics[width=\linewidth]{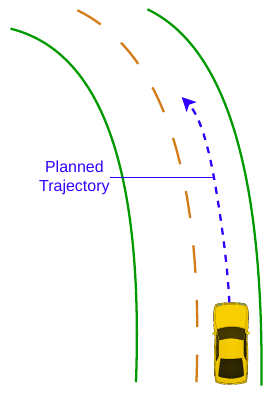}
      \caption{}
      \label{fig:motivation_1}
    \end{subfigure}\hfill
    \begin{subfigure}[t]{0.33\columnwidth}
      \includegraphics[width=\linewidth]{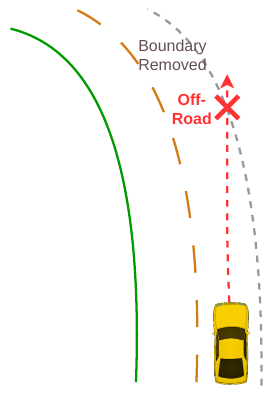}
      \caption{}
      \label{fig:motivation_2}
    \end{subfigure}\hfill
    \begin{subfigure}[t]{0.33\columnwidth}
      \includegraphics[width=\linewidth]{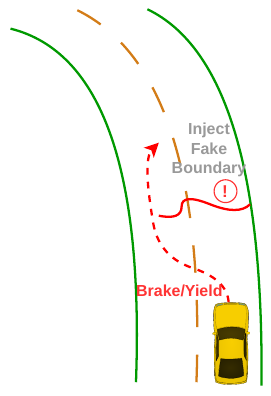}
      \caption{}
      \label{fig:motivation_3}
    \end{subfigure}
  \caption{\textbf{Motivating examples of mapping failures.} (a) Benign. (b) Boundary removal. (c) Boundary injection.}
  \label{fig:motivation}
\end{figure}

Camera-based online HD map construction is a safety-critical perception component of modern autonomous driving systems (ADS)~\cite{chen2024end}. These models ingest multi-view camera images and predict vectorized map elements---lane dividers, road boundaries, and pedestrian crossings---in bird's-eye view (BEV). The predictions directly feed downstream planning modules, where incorrect or missing map elements can cause lane violations, erroneous route decisions, or collisions~\cite{sato2025wip,zhang2025sok}, as shown in Figure~\ref{fig:motivation}.

The adversarial robustness of perception models has been widely studied in the context of $L_p$-bounded pixel perturbations~\cite{goodfellow2015explaining, madry2018towards, carlini2017towards}, physical-world patches~\cite{eykholt2018robust, cao2019adversarial,sato2021dirty}, and recently, adversarial patches targeting HD map construction specifically~\cite{lou2025asymmetry}. However, these attacks share a fundamental limitation from a security standpoint: the adversarial signal is either additive pixel noise or localized patches, producing high-frequency artifacts that are detectable and removable by standard adversarial defenses\cite{dong2020benchmarking,cohen2019certified, nie2022diffpure}. A defender deploying even lightweight preprocessing such as JPEG compression~\cite{guo2017countering} can substantially mitigate the threat.

Nevertheless, previous experimental security research on Tesla's AutoPilot~\cite{tencent2019experimental} showed that its HD mapping model could be misled by heuristic appearance changes on road marking (\eg serrated lane). This suggests semantic-level changes can be a source of vulnerability of mapping models. 
However, systematic discovery of such semantic mutations leading to perception failures remain open, as the direct optimization of pixel-level perturbations in classic adversarial attacks does not translate into semantic-level changes. 

To optimize and search in the abstract semantic space, diffusion-based driving scene generators~\cite{rombach2022high,gao2023magicdrive,wang2024drivedreamer} create a potential attack surface fundamentally different than pixel-level perturbation. Such diffusion models follow the design philosophy of Variational Auto-Encoders (VAEs)~\cite{kingma2013vae}, which models the latent distribution of real-world driving data as a Gaussian and learns a denoising network transforming randomly sampled latents into diverse photo-realistic multi-view driving images. Since these generators are trained on large-scale real-world driving data, their latent manifold encodes a broad distribution of plausible environmental variation---shadow patterns, wet road surfaces, surface textures~\cite{mokady2023null}, which may capture natural challenging cases for perception models\footnote{Examples in Figures~\ref{fig:nuscenes_error} and~\ref{fig:rw_error} in Appendix.}.

However, leveraging such diffusion models face challenges in faithfulness and controllability. Specifically, the driving scene generators are engineered for diversity in generated samples. Directly applying random generation or text-prompt guided generation alone is not enough due to loose correlation with the ground truth scene and non-actionable semantic changes. For example, the generated images may correspond to an entirely new location (\eg with different background fixtures or weathers) and/or different actual road topology (\eg changing a two-lane drive into one) instead of controlled semantic changes on a given sample (\eg specific patterns on the road the attacker can apply), as shown in Figure~\ref{fig:challenges}. 

\begin{figure}[t]
  \centering
\includegraphics[width=\columnwidth]{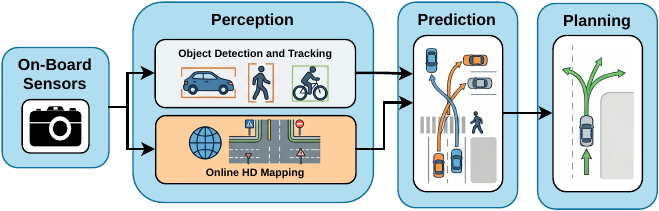}
  \caption{Online HD mapping enables the autonomous vehicles to perceive surrounding road layout, which is essential for driving safety.}
  \label{fig:av_system}
\end{figure}

We present \sysname, a semantic adversarial attack exploration framework that exploits the latent space structure of diffusion models. \sysname systematically searches in the latent manifold to identify legitimate samples from the model's learned distribution of real-world driving scenes with no additive pixel-level perturbation yet leading to perception failures. 
To address the challenges of faithfulness and controllability, \sysname $(1)$ \textit{Invert the ground truth images} into the latent space as the anchor point, avoiding generation of irrelevant scenes; $(2)$ Leverage  ControlNet~\cite{zhang2023adding} to constraint the generation to be \textit{conditioned on the ground truth mapping results}, ensuring the actual road topology remains consistent; $(3)$ Use direction loss on CLIP~\cite{radford2021learning} to \textit{guide the search} toward specific semantic changes (\eg wet road) that the attacker can influence.

\noindent We demonstrate two complementary attack goals with \sysname:
\begin{itemize}
    \item \textbf{Boundary removal}: Suppress detection of real road boundaries so that the planner routes through lane markings into opposing traffic or off-road.
    \item \textbf{Boundary injection}: Encourage detection of fictitious road boundaries at attacker-chosen positions, forcing brake or swerve to avoid an obstacle that does not exist. 
\end{itemize}

We comprehensively evaluate \sysname on nuScenes~\cite{caesar2020nuscenes} against the state-of-the-art HD mapping model MapTR~\cite{liao2023maptr}, with pixel PGD~\cite{madry2018towards} and AdvPatch~\cite{lou2025asymmetry} (CCS\,'25) as baselines. On \emph{attack effectiveness}, \sysname suppresses 57.7\% of road-boundary detections (close to the white-box upper bound of pixel PGD at 72\%) and is the \emph{only} method we evaluate that reliably injects fictitious boundaries (+1.88 detections per scene), translating into a 33\% planner false-stop rate and up to 52\% off-road rate. On \emph{robustness against defenses}, three standard input-preprocessing defenses (JPEG compression~\cite{dziugaite2016study}, median filtering~\cite{guo2017countering}, and DiffPure~\cite{nie2022diffpure}) recover 54--81\% of pixel PGD's suppressed detections but only 19--36\% of \sysname's. Defenses tuned to high-frequency $L_p$ noise do not generalize to semantic perturbations drawn from the data distribution.

For \emph{realism quantification}, two independent VLM judges (InternVL3-8B~\cite{internvl3}, Gemma-4-E4B~\cite{gemma4}) classify \sysname's adversarial samples as realistic 80--84\% of the time---within 14--17\% of clean nuScenes (96--98\%) and far above pixel PGD (28--52\%) and AdvPatch (0--9\%). For \emph{physical realizability}, we conduct a real-world proof-of-concept case study on a closed road in which team members hand-reproduced the dominant structural cues of \sysname's optimized patterns in sidewalk chalk and re-collected the data through the same camera rig. Even this deliberately low-fidelity, low-cost reproduction---which captures only a coarse outline and ignores detailed texture and edge sharpness---is sufficient to suppress or injection road element detections in the same direction \sysname predicted, suggesting that the discovered vulnerabilities survive the digital-to-physical gap with substantial tolerance to implementation error.

\noindent Our contributions are as follows:
\begin{itemize}
    \item We introduce \sysname, a framework for systematic discovery of \textbf{semantic attacks} against online HD mapping that searches 
    the latent space of a diffusion model trained on real driving data, discovering physically plausible semantic changes leading to perception failure, without pixel-level noise (\S\ref{sec:attack}). Code is available at \url{https://anonymous.4open.science/r/MIRAGE-F7A9/} for public use and validation.
    \item We evaluate \sysname on nuScenes with two attack goals---\textbf{boundary removal} (57.7\% detection suppression) and \textbf{boundary injection} (the only method that successfully injects fictitious boundaries) (\S\ref{sec:effectiveness}).
    \item We conduct a systematic \textbf{defense evasion evaluation} across three defense families (JPEG~\cite{dziugaite2016study}, median filtering~\cite{guo2017countering}, DiffPure~\cite{nie2022diffpure}) and show that median filtering recovers 80.7\% of pixel PGD's suppressed detections but only 35.7\% of \sysname's. \sysname's detection suppression also survives DiffPure purification (32.6\% recovery) while pixel PGD's is largely neutralized (80.4\% recovery).
    \item We validate \textbf{perceptual realism} on 1,000 images across 5 categories using two independent VLM judges: \sysname passes as realistic 80--84\% of the time versus 28--52\% for pixel PGD and 0--9\% for AdvPatch. We further show that \sysname's discovered failure modes appear spontaneously in the original nuScenes dataset and real-world footage we collected (\S\ref{sec:realism}).
    \item We conduct a \textbf{real-world proof-of-concept case study} demonstrating even a low-fidelity, low-cost physical reproduction of \sysname's optimized patterns is sufficient to mislead mapping model's predictions in the direction the attacker intended (\S\ref{sec:physical_feasibility}).
\end{itemize}

%% file: secs/related_work.tex
\section{Related Work}
\label{sec:background}

\subsection{Perception in Autonomous Driving Systems}
\label{sec:bg_perception}

Modern autonomous driving systems (ADS), either end-to-end or explicitly modularized, decompose environmental understanding into a stack of perception models whose outputs feed a downstream prediction--planning pipeline~\cite{hu2023uniad}, as shown in Figure~\ref{fig:av_system}. The perception layer ingests raw sensor streams---most commonly surround-view cameras, complemented by LiDAR or radar---and produces a structured, machine-readable representation of the driving scene in bird's-eye view (BEV)~\cite{hu2023uniad, jiang2023vad}. Because every subsequent decision is made on top of this representation, any error introduced at the perception stage propagates directly into motion forecasting and trajectory planning, which makes perception both a primary target for adversaries and a central concern for ADS safety~\cite{zhang2025sok}.

Two perception sub-tasks are load-bearing for safe driving: object detection/tracking and online HD map construction:

\noindent\textbf{Object detection and tracking} reasons about the \emph{dynamic} scene. It localizes traffic participants (\eg vehicles, pedestrians, cyclists), estimates their attributes (class, size, velocity, heading), and assigns each instance a persistent identity across frames~\cite{Jia2020TrackerHijack, wang2023does, muller2022physical, wang2024physical}. These outputs are the direct input to the motion-prediction module, which predicts each tracked object's short-horizon trajectory so that the ego vehicle can anticipate interactions such as cut-ins and crossings. Perception failures such as missed detection, spurious objects, or identity switches can translate into wrong forecasts, which in turn cause unsafe driving behaviors~\cite{jia2019fooling,zhang2025sok}.

\noindent\textbf{Online HD map construction}, by contrast, reasons about the \emph{static} scene. Given $N$ calibrated surround-view camera images with known intrinsic and extrinsic parameters, a map construction model $F$ produces vectorized map elements---lane dividers, road boundaries, and pedestrian crossings---in BEV coordinates~\cite{gao2020vectornet}: $
\hat{\mathbf{y}} = F(\mathbf{x}_1, \ldots, \mathbf{x}_N),
$
where each $\hat{\mathbf{y}}_i$ is a polyline with class label and confidence score~\cite{gao2020vectornet, liu2023vectormapnet}. The canonical model  MapTR~\cite{liao2023maptr, liao2024maptrv2} formulates map construction as a set prediction problem using a transformer decoder with learnable queries and has been widely incorporated into state-of-the-art end-to-end driving stacks~\cite{jiang2023vad, sun2025sparsedrive, gao2025rad, zheng2024genad}. Unlike classical pipelines that depend on pre-built offline HD maps~\cite{sato2025wip}, online construction enables the AV to operate in regions where pre-mapped data is unavailable, stale, or incorrect, and supplies the planner with the drivable region, lane topology, and legal driving corridors required to select feasible trajectories~\cite{lou2025asymmetry}.

Downstream prediction and planning consume these perception results: the motion predictor conditions dynamic-agent forecasts on the road structure---lanes constrain plausible future paths---while the planner fuses predicted agent trajectories with the map to select a safe and rule-compliant route~\cite{hu2023uniad, jiang2023vad}. A planner that does not see a road boundary may route across it, and a planner that sees a fictitious boundary may halt or swerve unnecessarily~\cite{zhang2025sok}. 
This tight coupling is what renders perception---and, as we show, online HD mapping in particular---a consequential attack surface~\cite{lou2025asymmetry,sato2025wip}.

\subsection{Adversarial Attacks on Perception}
\label{sec:bg_attacks}

Perception systems in ADS have been shown vulnerable to a broad spectrum of adversarial inputs. Digital attacks follow the standard $L_p$-bounded paradigm~\cite{goodfellow2015explaining, madry2018towards, carlini2017towards}; ADS-specific instantiations have targeted LiDAR point clouds~\cite{cao2019adversarial, sun2020towards, tu2020physically}, camera-based classification~\cite{eykholt2018robust}, and multi-object detection and tracking~\cite{Jia2020TrackerHijack, wang2023does, jia2019fooling}. Physical-world attack vectors include printed patches~\cite{wei2024physical, eykholt2018robust}, projected light patterns and targeted illumination~\cite{muller2022physical, muller2025investigating,lou2025asymmetry}, and adversarial driving maneuvers by neighboring agents~\cite{yamanaka2020adversarial,song2023discovering, wang2024physical}. 
Also, the work by~\cite{sato2021dirty} demonstrated that printed road patches displaying adversarial perturbations can affect the automated lane-centering models, whereas researchers at Tencent~\cite{tencent2019experimental} showed Tesla AutoPilot's lane detection algorithms can be fooled by irregularly shaped lane markings. 
Most recently, Lou et al.~\cite{lou2025asymmetry} demonstrated the first white-box adversarial patch (AdvPatch) targeting online HD map construction, optimizing a learnable roadside patch via differentiable 3D-to-2D projection. While effective, AdvPatch, like $L_p$-bounded and patch-based methods in general, introduces localized high-frequency artifacts that are detectable and removable by standard input preprocessing~\cite{guo2017countering}.

A growing line of work uses generative models as the \emph{source} of adversarial perturbations. Yuan et al.~\cite{yuan2023precise} leverage regularized GANs to generate semantically-mutated images and develop a certified robustness framework applicable in the semantic space of GANs. Liang et al.~\cite{liang2023adversarial} and Zhuang et al.~\cite{zhuang2023pilot} craft adversarial perturbations \emph{against} diffusion-based image generation, targeting generation quality rather than downstream perception. The work by~\cite{xue2023diffusion,dai2024advdiff} use diffusion model to generate images that fool classification models. Other work~\cite{xie2024advdiffuser, xu2025diffscene} leverage guided diffusion to generate safety-critical trajectory patterns for background vehicles. Sato et al.~\cite{sato2024intriguing} use a fuzzing approach on text-prompt guidance of diffusion model to generate adversarial patterns that are recognized as stop signs by object detectors while appearing irrelevant to humans. \sysname departs from these prior work by searching the \emph{latent manifold} of a multi-view driving scene diffusion model. 
The diffusion model serves as a \emph{search tool} over the space of plausible environmental conditions; the actual security threat is the existence of naturally occurring configurations---specific shadow angles, road texture patterns---that reliably degrade HD map perception at chosen locations while the actual road topologies remain the same.\looseness=-1

\subsection{Adversarial Defenses}
\label{sec:bg_defenses}

Defenses against adversarial examples aim at removing or neutralizing the perturbation before it reaches the model: JPEG compression~\cite{dziugaite2016study}, Gaussian smoothing~\cite{cohen2019certified}, median filtering~\cite{guo2017countering}, and, most recently, diffusion-based purification (DiffPure)~\cite{nie2022diffpure}, which projects an input back onto the natural image distribution by running a short forward-reverse diffusion pass. All preprocessing defenses share the same implicit premise---that adversarial perturbations are high-frequency, spatially localized, or otherwise statistically distant with respect to the natural image distribution. \sysname is designed from the outset to violate this premise: its outputs are sampled from a diffusion prior learned on real driving scenes with no out-of-distribution signal for preprocessing to strip away. We quantify this effect across three defense configurations in \S\ref{sec:defense}.

%% file: secs/threat_model.tex
\section{Threat Model}
\label{sec:threat}

We consider an autonomous vehicle (AV) that relies on a camera-based online mapping model to construct HD map elements---road boundaries, lane dividers, and pedestrian crossings---from surround-view images. These map predictions feed directly into a local motion planner that generates trajectories. This architecture is representative of modern autonomous driving stacks~\cite{jiang2023vad,sun2025sparsedrive,gao2025rad,zheng2024genad}.

\begin{figure*}[t]
  \centering
  \begin{subfigure}[t]{0.23\textwidth}
      \includegraphics[width=\linewidth]{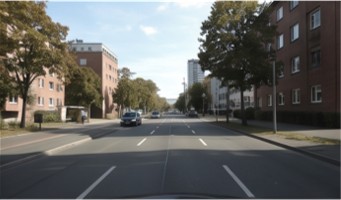}
      \caption{Clean}
      \label{fig:challenge_clean}
    \end{subfigure}\hfill
    \begin{subfigure}[t]{0.23\textwidth}
      \includegraphics[width=\linewidth]{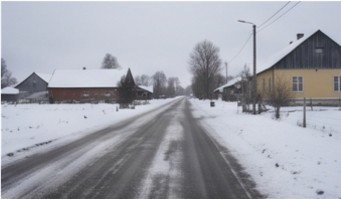}
      \caption{Irrelevant}
      \label{fig:challenge_random}
    \end{subfigure}\hfill
    \begin{subfigure}[t]{0.23\textwidth}
      \includegraphics[width=\linewidth]{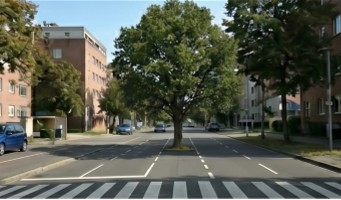}
      \caption{Unrealizable}
      \label{fig:challenge_unactionable}
    \end{subfigure}\hfill
    \begin{subfigure}[t]{0.23\textwidth}
      \includegraphics[width=\linewidth]{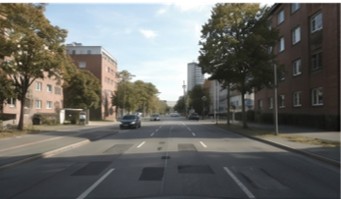}
      \caption{Plausible}
      \label{fig:challenge_mirage}
    \end{subfigure}
  \caption{Illustration of the challenges in faithfulness and controllability of semantic changes. (a) Clean reference input. (b) Random generation yields irrelevant scenes. (c) Unconditioned and non-CLIP guided generation results in changes of actual lane markings, and/or unrealizable mutations. (d) \sysname generates faithful scene via latent inversion and controllable semantic mutations (road appearance) with CLIP direction guidance.}
  \label{fig:challenges}
\end{figure*}

\subsection{Attacker Goals}
\label{sec:attacker}

The attacker's objective is to discover specific environmental conditions, such as shadow patterns and wet road surfaces, that cause the mapping model to produce incorrect predictions at inference time. We consider two concrete attack goals:
\begin{itemize}
    \item \textbf{Boundary removal}: Suppress detection of real map elements (dividers, road boundaries), enabling the planner to route into opposing traffic or off-road.
    \item \textbf{Boundary injection}: Fabricate fictitious road boundaries at attacker-chosen positions, forcing the planner to perceive impassable barriers and trigger emergency braking or unsafe lane changes.
\end{itemize}

\subsection{Attacker Capabilities}
\label{sec:capabilities}

We assume the attacker has access to a conditional diffusion-based driving scene generator (\S\ref{sec:prelim}) and white-box access to the victim HD mapping model. White-box access is the standard upper-bound setting in adversarial ML security evaluation~\cite{carlini2017towards}; it is realistic here because (a) production-grade perception models are increasingly trained on top of openly-published architectures and pretrained checkpoints~\cite{he2016deep}; 
and (b) AV perception model developers and red teams can leverage \sysname to systematically discover vulnerabilities of the developing model for investigation and improvement. 

The diffusion model is used solely as a \emph{vulnerability discovery tool}: by searching the model's latent space---which encodes the full range of real-world environmental variation observed in its training data---the attacker systematically identifies which semantic conditions maximize perception failure. For deployment, an attacker can physically reproduces plausible semantic conditions discovered by \sysname in the real-world (\eg a shadow projection, tainted road). We present a physical realizability case study in \S\ref{sec:physical_feasibility}.\looseness=-1

\paragraph{Planner Assumed Black-box.} Although our attack objective is differentiated through the mapping model, we make \emph{no} assumption of access to the downstream planner. The attacker does not query the planner during optimization, does not differentiate through its trajectory search or policy, and does not require any knowledge of its cost functions or hyperparameters. Following the prior art~\cite{lou2025asymmetry}, our formal safety evaluation runs a classical A*\ planner on the HD mapping's BEV output (\S\ref{sec:effectiveness}). We additionally include a transferability case study against the end-to-end VAD planner~\cite{jiang2023vad} (\S\ref{sec:transfer}), whose weights we never access during attack optimization: adversarial samples crafted solely from the HD mapping model gradients are fed zero-shot to VAD. This strengthens the realism of the threat model---a capability restricted to mapping-model gradients still propagates to trajectory-level harm.

%% file: secs/methodology.tex
\section{Design of \sysname Framework}
\label{sec:method}

\sysname reframes adversarial example generation as a constrained search over the latent space of a diffusion-based driving scene generator. Realizing this idea requires solving three challenges of faithfulness and controllability that naive use of a pretrained generator does not address, as shown in Figure~\ref{fig:challenges}: 
\texttt{(C1)}~Generation without a latent anchor yields scenes unrelated to the original input, which we prevent by inverting the ground-truth images into encoded latents at initialization and re-rendering through only a partial diffusion interval; \texttt{(C2)}~Unconditional sampling can alter the underlying road topology (\eg changing number of lanes), which we prevent by enforcing the ground-truth BEV layout as a ControlNet~\cite{zhang2023adding} conditioning signal;  and \texttt{(C3)}~Unconstrained latent search can drift toward unrealizable changes such as warped buildings or sky, which we address by introducing CLIP~\cite{radford2021learning} direction loss that steers edits toward targeted environmental variation. The remainder of this section makes these design choices precise.

\subsection{Preliminaries}
\label{sec:prelim}

We first introduce two technical foundations that \sysname builds upon: conditional latent diffusion models, which provide the generative prior over driving scenes, and CLIP-based direction losses, which steer edits toward named semantic changes. We summarize the aspects of each that are essential to the method.

\subsubsection{Latent Diffusion Models.}
Diffusion models~\cite{ho2020denoising} generate data by iteratively denoising a sample from a Gaussian prior. A \emph{latent diffusion model} (LDM)~\cite{rombach2022high} operates in a compressed latent space produced by a variational autoencoder (VAE)~\cite{kingma2013vae}: an encoder $\mathcal{E}$ maps an image $\mathbf{x}$ to a latent code $\mathbf{z}_0 = \mathcal{E}(\mathbf{x})$, and a decoder $\mathcal{D}$ reconstructs $\hat{\mathbf{x}} = \mathcal{D}(\mathbf{z}_0)$. During training, the forward process adds Gaussian noise to $\mathbf{z}_0$ over $T$ timesteps according to a variance schedule $\{\beta_t\}_{t=1}^T$:
\begin{equation}
q(\mathbf{z}_t \mid \mathbf{z}_0) = \mathcal{N}\!\left(\mathbf{z}_t;\; \sqrt{\bar{\alpha}_t}\,\mathbf{z}_0,\; (1 - \bar{\alpha}_t)\,\mathbf{I}\right),
\label{eq:forward}
\end{equation}
with $\alpha_t = 1 - \beta_t$ and $\bar{\alpha}_t = \prod_{s=1}^t \alpha_s$. A denoising network $\epsilon_\theta$ (typically a U-Net~\cite{ronneberger2015unet}) learns to predict the noise component given $\mathbf{z}_t$ and $t$. In \emph{conditional} LDMs, $\epsilon_\theta$ additionally ingests a text prompt embedding $\mathbf{e}$ (injected via cross-attention) and spatial conditioning (injected via a ControlNet~\cite{zhang2023adding} branch that adds residual features to the U-Net). At inference, the deterministic DDIM sampler~\cite{song2021denoising}
\begin{equation}
\mathbf{z}_{t-1} = \sqrt{\bar{\alpha}_{t-1}}\, \hat{\mathbf{z}}_0(t) + \sqrt{1 - \bar{\alpha}_{t-1}}\, \boldsymbol{\epsilon}_\theta(\mathbf{z}_t, t),
\label{eq:ddim}
\end{equation}
with $\hat{\mathbf{z}}_0(t) = \bigl(\mathbf{z}_t - \sqrt{1-\bar{\alpha}_t}\,\boldsymbol{\epsilon}_\theta(\mathbf{z}_t,t)\bigr)/\sqrt{\bar{\alpha}_t}$, maps noise to images in a small number of steps and admits \emph{inversion}: given a clean latent $\mathbf{z}_0$, one can recover the noise $\mathbf{z}_T$ that would reconstruct it~\cite{mokady2023null}, yielding a differentiable, nearly bijective map between latent codes and generated images. A ControlNet~\cite{zhang2023adding} can be added to enforce the generated output respect particular conditions such as text-prompt or camera projection matrices~\cite{gao2023magicdrive}.

\subsubsection{CLIP-based Semantic Guidance.}
Contrastive Language-Image Pre-training (CLIP)~\cite{radford2021learning} jointly embeds images and text into a shared latent space in which semantic similarity is measured by cosine distance. StyleGAN-NADA~\cite{gal2022stylegan} and DiffusionCLIP~\cite{kim2022diffusionclip} leverage this alignment to guide image edits toward text-specified semantic changes by optimizing a \emph{direction loss}: the cosine similarity between the direction of change in the image embedding and a target text direction in CLIP space. We adapt this mechanism (\S\ref{sec:clip}) to constrain adversarial latent perturbations toward physically realizable environmental changes---shadows, wet surfaces, surface texture---thereby ensuring the discovered vulnerabilities correspond to actionable and realizable conditions.

\begin{figure}[t]
  \centering
  \includegraphics[width=\columnwidth]{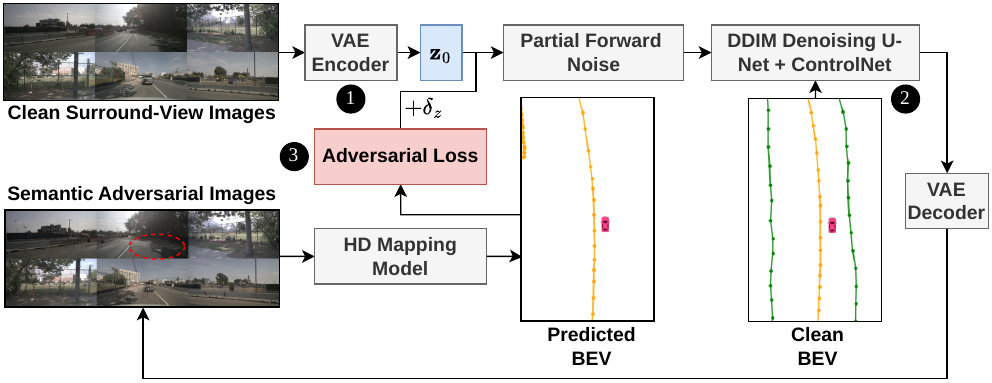}
  \caption{\textbf{\sysname pipeline.} Given a driving scene, \sysname inverts the surround-view images into per-view latent codes $\mathbf{z}_0$, then searches for a nearby latent $\mathbf{z}_0 + \boldsymbol{\delta}_z$ that, when decoded through the conditional diffusion model (BEV ControlNet), produces images degrading the mapping model's predictions while having the same road topology.}
  \Description{Pipeline diagram showing latent inversion, perturbation search in latent space, conditional diffusion decoding, and mapping model attack.}
  \label{fig:pipeline}
\end{figure}

\subsection{Overview}
\label{sec:overview}

\sysname operates in three stages, as shown in Figure~\ref{fig:pipeline}. \circled{1} We \emph{invert} the ground-truth driving images into the diffusion model's latent space to obtain the latent code $\mathbf{z}_0$ that reconstructs them, establishing a starting anchor within the semantic space of real-world scenes. Note that in the benign operation of driving scene generation models, $\mathbf{z}_0$ is randomly sampled from a Gaussian prior to provide scene diversity. The inversion is required to address \texttt{(C1)}. \circled{2} We run forward pass and \emph{search} for a nearby latent $\mathbf{z}_0 + \boldsymbol{\delta}_z$ with clean BEV conditioned ControlNet~\cite{zhang2023adding} such that the resulting images adhere to the original road topology without structural changes (which the attacker cannot manipulate), addressing \texttt{(C2)}. \circled{3} By optimizing the adversarial loss, \sysname identifies perturbed latent codes corresponding to semantic changes in the generated views that can maximally degrade the mapping  predictions. Here, a CLIP direction loss constrains the search toward physically plausible semantic changes, steering perturbations onto realizable road-surface variation (shadows, texture) rather than arbitrary mutations, addressing \texttt{(C3)}. The whole \sysname pipeline is described in Algorithm~\ref{alg:mirage}.

\input{secs/alg_mirage}

Since the diffusion model was trained on real driving imagery, its latent space encodes the natural distribution of environmental conditions. Nearby latents correspond to the same scene under different but plausible conditions---varying shadows, moisture, surface texture. \sysname exploits this structure: rather than adding arbitrary pixel noise, it navigates within the manifold of realistic scenes to find conditions that happen to degrade perception.

\subsection{Latent Inversion and Perturbation}
\label{sec:attack}

Given ground-truth driving images $\mathbf{x}_{\text{gt}} = \{\mathbf{x}_i\}_{i=1}^N$ from $N$ surround-view cameras, we first encode each view independently into the VAE latent space to obtain per-view latents $\mathbf{z}_0^{(i)} = \mathcal{E}(\mathbf{x}_i)$, where $\mathcal{E}$ is the encoder of the diffusion model's VAE (\S\ref{sec:prelim}). For brevity we write $\mathbf{z}_0$ for the stacked $N$-view latent in equations below; adversarial perturbation operates on a per-view latent codes $\boldsymbol{\delta}_z^{(i)}$ for each camera. Cross-view consistency is not enforced through a shared latent but through the \emph{shared conditioning} $(\mathbf{m}, \mathbf{e}, \mathbf{c})$---the clean BEV map $\mathbf{m}$ specifies a single road topology seen from every camera, the text prompt $\mathbf{e}$ fixes the global scene description, and per-camera projection parameters $\mathbf{c}$ tie each view to its intrinsics and extrinsics. This factorization matches surrounding-view driving scene generator's~\cite{gao2023magicdrive, wang2024drivedreamer} pretraining regime and is what allows partial-diffusion edits to remain 3D-consistent at the topology level: the clean BEV-conditioned road structure is identical before and after perturbation. We do not explicitly enforce 3D consistency of the \emph{edits} themselves (\eg a shadow rendered in \texttt{CAM\_FRONT} need not project consistently into \texttt{CAM\_FRONT\_LEFT}); empirically the joint mapping-model loss produces view-coherent changes because gradients from the shared BEV prediction flow back through every view.\looseness=-1

To generate a modified image from a perturbed latent while preserving overall scene structure, we adopt a partial-diffusion strategy. We add noise to the perturbed latent at a partial timestep $t^* = \lfloor s \cdot T \rfloor$, determined by a strength parameter $s \in (0, 1)$ that controls how much of the original scene structure is preserved:
\begin{equation}
\mathbf{z}_{\text{noisy}} = \sqrt{\bar{\alpha}_{t^*}}\, (\mathbf{z}_0 + \boldsymbol{\delta}_z) + \sqrt{1 - \bar{\alpha}_{t^*}}\, \boldsymbol{\epsilon},
\label{eq:noise}
\end{equation}
where $\boldsymbol{\epsilon}$ is randomly sampled from $\mathcal{N}(\mathbf{0}, \mathbf{I})$ and fixed across PGD iterations for gradient stability (without this, stochastic noise resampling at every step produces a moving optimization target and destabilizes PGD). $\bar{\alpha}_{t^*}$ is the cumulative noise schedule coefficient at timestep $t^*$, and $\boldsymbol{\delta}_z$ is a \emph{per-view} adversarial perturbation bounded by $\|\boldsymbol{\delta}_z\|_\infty \leq \epsilon$. Although $\boldsymbol{\delta}_z$ is view-local, the joint optimization against the mapping model indirectly couples the perturbations: gradients from the shared BEV prediction flow back into every view, so successful suppression or injection requires view-coherent semantic change. 
The noisy latent is then denoised through $K$ DDIM steps (Eq.~\ref{eq:ddim}), conditioned on the BEV map $\mathbf{m}$, text prompt embedding $\mathbf{e}$, and camera parameters $\mathbf{c}$ via ControlNet~\cite{zhang2023adding}:
\begin{equation}
\mathbf{x}_{\text{adv}} = \mathcal{D}\!\left(\text{DDIM}(\mathbf{z}_{\text{noisy}},\, \mathbf{e},\, \mathbf{m},\, \mathbf{c})\right),
\end{equation}\label{eq:x_adv}
where $\mathcal{D}$ is the VAE decoder. 
We run only the \emph{last} $t^*$ timesteps of the denoising schedule---in effect, inverting forward to noise level $t^*$, perturbing, and then denoising back through that same interval. A small strength ($s{=}0.3$, \eg last 6 of 20 DDIM steps) is chosen deliberately: larger $s$ gives the denoiser more freedom and yields bolder edits but drifts away from the original scene layout (vehicle positions, building facades, sky); smaller $s$ keeps the output near the ground truth but allows too little room for semantic change\footnote{The $s{=}0.3$ point empirically preserves scene structure while leaving enough generative freedom to re-render road-surface details (texture, shadows) under $\boldsymbol{\delta}_z$. A full sweep over $s$ is orthogonal to our other ablations and is noted as future work.}.

\paragraph{Why Latent Perturbation.}
The conditioning interface (Eq.~\ref{eq:x_adv}) exposes multiple potential attack surfaces: the starting latent $\mathbf{z}_0$, the text prompt embedding $\mathbf{e}$, and the BEV control map $\mathbf{m}$. We explored prompt-embedding perturbation ($\boldsymbol{\delta}_e$ on the text encoder output) in preliminary experiments; it tended to alter the whole scene globally rather than modify road-surface semantics, and is not reported further. Latent perturbation provides stronger gradients and more direct influence over generated image content, which we use exclusively in all experiments.

\subsection{Attack Objectives}
\label{sec:objectives}

\subsubsection{Boundary Removal.}
To suppress map element detections, we optimize the latent space perturbation $\boldsymbol{\delta}_z$ via PGD~\cite{madry2018towards} to minimize:
\begin{equation}
\mathcal{L}_{\text{remove}} = \lambda_{\text{conf}} \cdot \underbrace{\frac{1}{Q}\sum_{q=1}^{Q} \max_c \sigma(\hat{s}_{q,c})}_{\text{confidence suppression}} - \lambda_{\text{spread}} \cdot \underbrace{\frac{1}{Q}\sum_{q=1}^{Q} w_q \cdot \text{Spread}(\hat{\mathbf{p}}_q)}_{\text{point displacement}},
\label{eq:attack_loss}
\end{equation}
where $Q$ is the number of decoder queries in the mapping model, $\hat{s}_{q,c}$ are the classification logits for query $q$ and map element class $c$ (divider, boundary, crossing), $\sigma$ is the sigmoid function, $w_q$ is a confidence-based weight prioritizing high-confidence detections, and $\text{Spread}(\hat{\mathbf{p}}_q) = \frac{1}{N_p}\sum_i \|\hat{\mathbf{p}}_{q,i} - \bar{\mathbf{p}}_q\|^2$ measures geometric collapse of the predicted polyline points. The loss weights $(\lambda_{\text{conf}}, \lambda_{\text{spread}})$ are distinct from the PGD step size $\alpha$ used in optimization (\S\ref{sec:experiments}). Minimizing $\mathcal{L}_{\text{remove}}$ drives the model's confidence below the detection threshold while scattering predicted points away from true map elements; the threshold itself is a post-hoc evaluation parameter and is never used in the gradient.

\subsubsection{Boundary Injection.}
To inject fictitious boundaries, we target a BEV position $y^*$ (\eg 6\,m ahead of the ego vehicle) and select the $K$ queries whose predicted centroids are closest to $y^*$:
\begin{equation}
\mathcal{L}_{\text{inject}} = \underbrace{\sum_{k=1}^{K} \text{BCE}(\hat{s}_{k,\text{boundary}}, 1)}_{\text{inject confidence}} + \underbrace{\sum_{k=1}^{K} \|\hat{\mathbf{p}}_k - \mathbf{p}^*\|^2}_{\text{place at target}} + \gamma \cdot \mathcal{L}_{\text{suppress}},
\end{equation}
where $\mathbf{p}^* \in \mathbb{R}^{N_p \times 2}$ defines a horizontal polyline at $y^*$ and $\mathcal{L}_{\text{suppress}}$ reduces competing detections. A downstream planner encountering this injected boundary perceives an impassable barrier.

\paragraph{End-to-End Differentiability.}
The entire pipeline---DDIM denoising, VAE decoding, and mapping model inference---is differentiable, enabling gradient flow from $\mathcal{L}$ back to $\boldsymbol{\delta}_z$. We track gradients through the last $K$ denoising steps and use gradient checkpointing for memory efficiency~\cite{chen2016training}.

\subsection{CLIP Direction Loss for Semantic Guidance}
\label{sec:clip}

Unconstrained latent perturbation can produce visual changes that do not correspond to physically realizable conditions---for example, warping the sky or mutating the facades of nearby buildings. To ensure \sysname discovers \emph{actionable} vulnerabilities---environmental conditions an attacker could reproduce in the real world---we introduce a CLIP-based~\cite{radford2021learning} direction loss that steers perturbations toward specific semantic changes.

Let $V(\cdot)$ denote the CLIP vision encoder and $T(\cdot)$ the text encoder. We define a target semantic direction from a text description and measure whether the image-space change aligns with it:
\begin{equation}
\mathbf{d}_{\text{target}} = T(\text{target prompt}), \qquad \mathbf{d}_{\text{img}} = V(\mathbf{x}_{\text{adv}}) - V(\mathbf{x}_{\text{gt}}).
\end{equation}
The CLIP direction loss encourages this alignment while penalizing unwanted changes via a negative anchor:
\begin{equation}
\mathcal{L}_{\text{CLIP}} = 1 - \cos(\hat{\mathbf{d}}_{\text{img}}, \hat{\mathbf{d}}_{\text{target}}) + \lambda_{\text{neg}} \cdot \max\!\left(0,\; \cos(\hat{\mathbf{d}}_{\text{img}}, \mathbf{n})\right),
\label{eq:clip}
\end{equation}
where $\hat{\mathbf{d}}$ denotes $L_2$-normalized vectors, $\mathbf{n} = T(\text{negative anchor})$, and $\lambda_{\text{neg}}{=}0.5$. The total attack loss combines the perception objective with semantic guidance: $\mathcal{L} = \mathcal{L}_{\text{remove}} + \lambda_{\text{CLIP}} \cdot \mathcal{L}_{\text{CLIP}}$.

\paragraph{Prompts Used in Experiments.} We use the CLIP ViT-L/14~\cite{radford2021learning} encoder to match the Stable Diffusion 1.5 text encoder. Target prompts are \emph{``a road with shadows''} and \emph{``a wet road surface''}; the negative anchor is \emph{``distorted cars, warped buildings, unnatural sky''}, which covers the three non-road regions we want to leave unchanged. This constraint restricts the vulnerability search to physically plausible environmental manipulations. As we show in \S\ref{sec:clip_ablation}, the CLIP constraint steers toward plausible changes \emph{at no cost to attack effectiveness}---and in some configurations (notably boundary injection) improves it by concentrating perturbation energy on road surfaces.

%% file: secs/alg_mirage.tex
\begin{algorithm}[t]
\caption{\sysname: latent-space semantic attack on HD map construction.}
\label{alg:mirage}
\begin{algorithmic}[1]
\REQUIRE Surround-view images $\{\mathbf{x}_i\}_{i=1}^N$; BEV layout $\mathbf{m}$; prompt embedding $\mathbf{e}$; camera parameters $\{\mathbf{c}^{(i)}\}$; diffusion VAE $(\mathcal{E}, \mathcal{D})$; denoiser $\epsilon_\theta$; victim map model $f$; strength $s$; PGD iterations $K$; step size $\eta$; bound $\epsilon$; CLIP weight $\lambda_{\text{CLIP}}$; target text direction $\Delta\mathbf{t}$
\ENSURE Adversarial surround-view images $\{\mathbf{x}_{\text{adv}}^{(i)}\}_{i=1}^N$
\STATE \textit{// Stage \circled{1}: latent inversion (C1)}
\FOR{$i = 1$ \TO $N$}
    \STATE $\mathbf{z}_0^{(i)} \gets \mathcal{E}(\mathbf{x}_i)$
    \STATE Sample and fix $\boldsymbol{\epsilon}^{(i)} \sim \mathcal{N}(\mathbf{0}, \mathbf{I})$
    \STATE $\boldsymbol{\delta}_z^{(i)} \gets \mathbf{0}$
\ENDFOR
\STATE $t^* \gets \lfloor s \cdot T \rfloor$
\STATE \textit{// Stages \circled{2}--\circled{3}: BEV-conditioned PGD (C2) with CLIP guidance (C3)}
\FOR{$k = 1$ \TO $K$}
    \FOR{$i = 1$ \TO $N$}
        \STATE $\mathbf{z}_{\text{noisy}}^{(i)} \gets \sqrt{\bar{\alpha}_{t^*}}\,(\mathbf{z}_0^{(i)} + \boldsymbol{\delta}_z^{(i)}) + \sqrt{1 - \bar{\alpha}_{t^*}}\,\boldsymbol{\epsilon}^{(i)}$
        \STATE $\tilde{\mathbf{z}}^{(i)} \gets \text{DDIM}(\mathbf{z}_{\text{noisy}}^{(i)}, \epsilon_\theta;\, \mathbf{e}, \mathbf{m}, \mathbf{c}^{(i)})$ \quad\textit{// last $t^*$ steps}
        \STATE $\mathbf{x}_{\text{adv}}^{(i)} \gets \mathcal{D}(\tilde{\mathbf{z}}^{(i)})$
    \ENDFOR
    \STATE $\hat{\mathbf{y}} \gets f\!\left(\mathbf{x}_{\text{adv}}^{(1)}, \ldots, \mathbf{x}_{\text{adv}}^{(N)};\, \{\mathbf{c}^{(i)}\}\right)$
    \STATE $\mathcal{L} \gets \mathcal{L}_{\text{adv}}(\hat{\mathbf{y}}) + \lambda_{\text{CLIP}}\, \mathcal{L}_{\text{CLIP}}\!\left(\{\mathbf{x}_i, \mathbf{x}_{\text{adv}}^{(i)}\}_{i=1}^N,\, \Delta\mathbf{t}\right)$
    \FOR{$i = 1$ \TO $N$}
        \STATE $\mathbf{g}^{(i)} \gets \nabla_{\boldsymbol{\delta}_z^{(i)}} \mathcal{L}$
        \STATE $\boldsymbol{\delta}_z^{(i)} \gets \Pi_{\|\cdot\|_\infty \leq \epsilon}\!\left[\boldsymbol{\delta}_z^{(i)} - \eta\, \operatorname{sign}(\mathbf{g}^{(i)})\right]$
    \ENDFOR
\ENDFOR
\RETURN $\{\mathbf{x}_{\text{adv}}^{(i)}\}_{i=1}^N$
\end{algorithmic}
\end{algorithm}

%% file: secs/evaluation.tex
\section{Evaluation}
\label{sec:experiments}

\begin{figure}[t]
  \centering
  \includegraphics[width=\columnwidth]{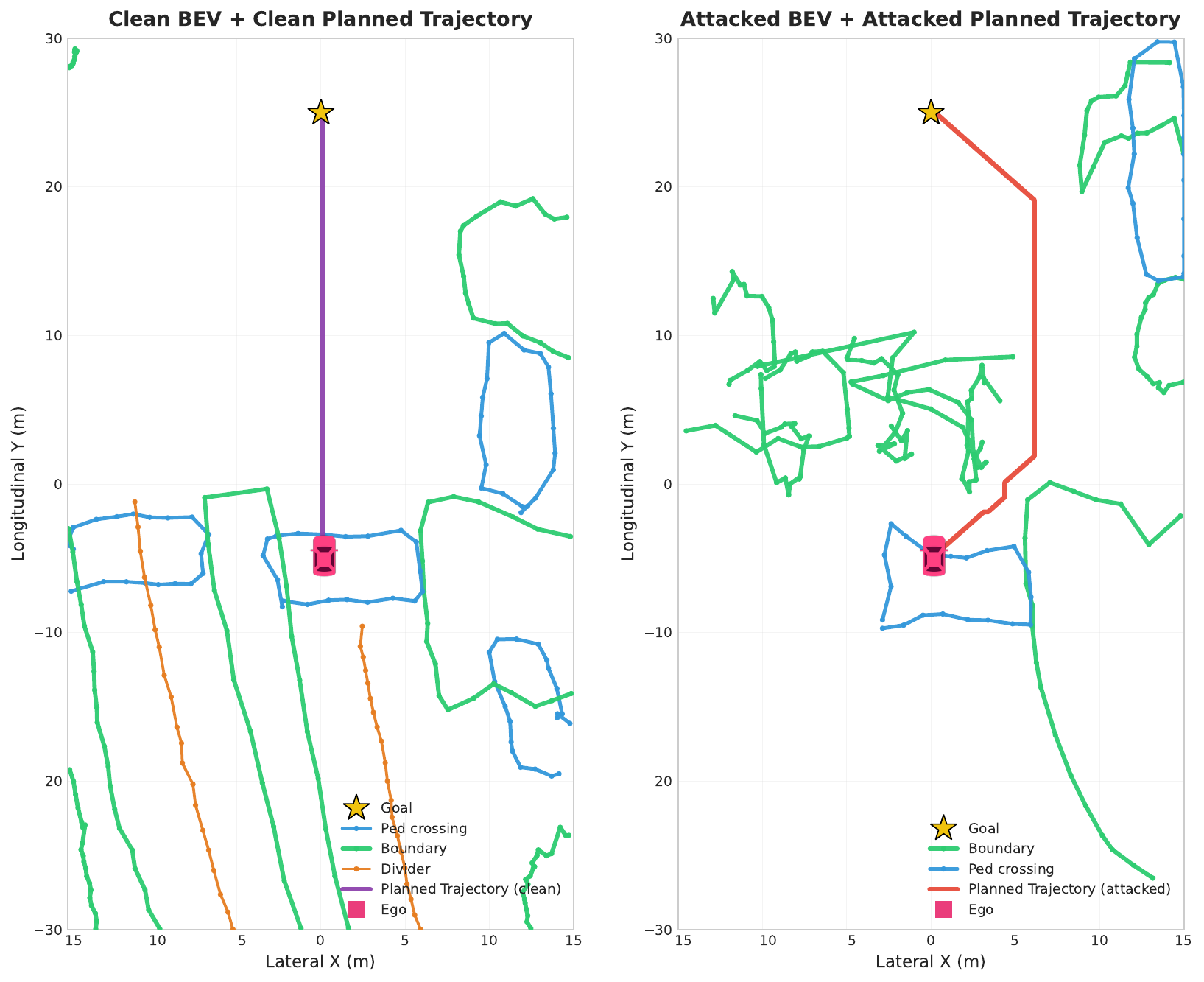}
  \caption{\textbf{Planner path corruption example.} Left: clean predictions with safe planning path. Right: the planner swerves to avoid a fictitious boundary.}
  \Description{BEV views showing clean planned path, adversarial path crossing into opposing lane, and path stopping at an injected boundary.}
  \label{fig:planner}
\end{figure}

\begin{figure*}[t]
  \centering
  \includegraphics[width=0.95\textwidth]{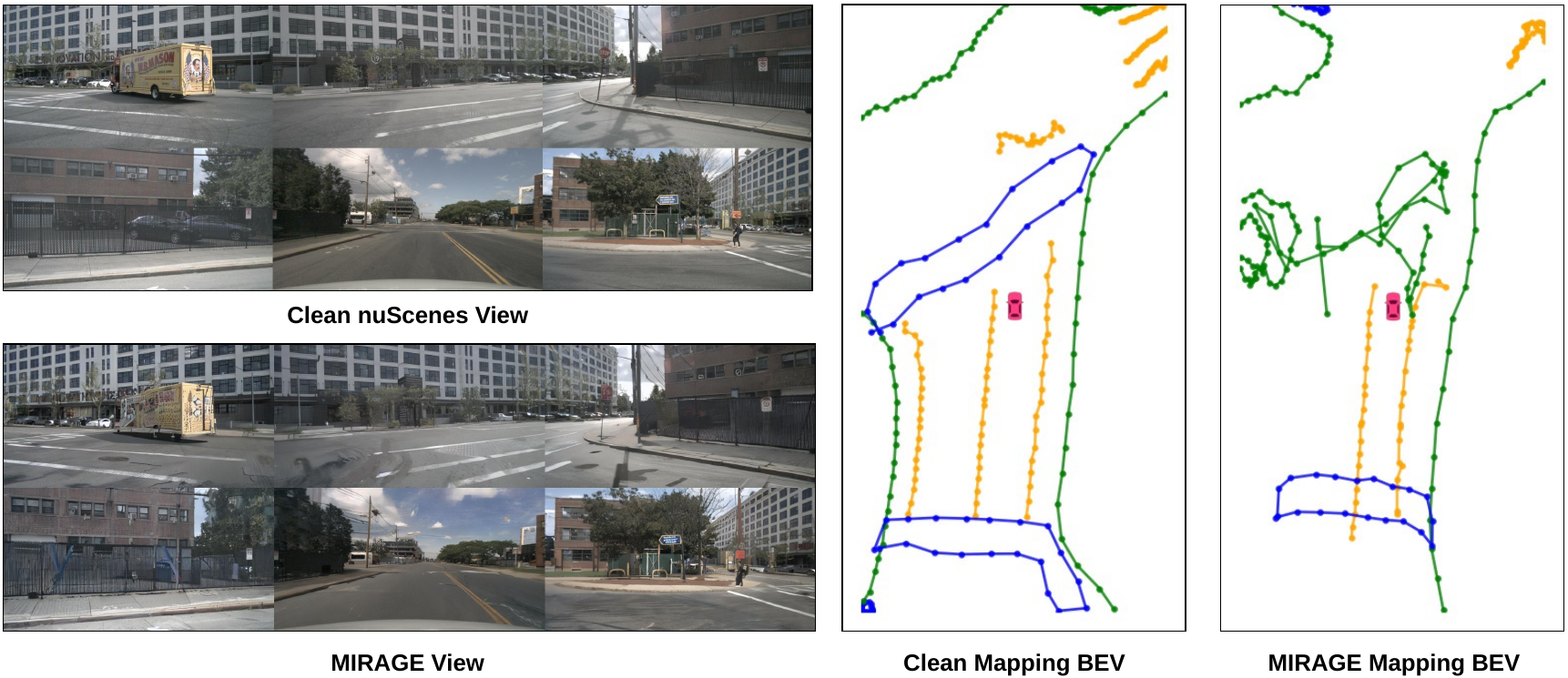}
  \caption{\textbf{Qualitative results.} Top-Left: clean 6-camera views from nuScenes. Bottom-Left: \sysname adversarial views ($\epsilon{=}0.5$, CLIP shadow direction). Middle: clean BEV mapping prediction. Right: attacked BEV predictions showing fictitious boundaries. 
  Additional qualitative samples including PGD and AdvPatch with VLM judgments are provided in Appendix~\ref{app:realism_prompt}.}
  \Description{Six-camera driving views showing clean images with map overlay, adversarial images with shadow-like changes, and clean/adv BEV predictions.}
  \label{fig:qualitative}
\end{figure*}

\subsection{Experimental Setup}

\noindent\textbf{Dataset and Models.} We evaluate on nuScenes~\cite{caesar2020nuscenes} with 6-camera surround views at 424$\times$800 resolution. For each of boundary injection and removal attacks, we randomly select 100 scenes from the validation set for evaluation. Following the prior art~\cite{lou2025asymmetry}, we use MapTR~\cite{liao2023maptr} as the victim mapping model, and an A* path planner operating on the mapping model's BEV predictions for downstream safety evaluation (\S\ref{sec:effectiveness}).
For the diffusion-based driving scene generator, we adopt MagicDrive~\cite{gao2023magicdrive} (Stable Diffusion 1.5~\cite{rombach2022high} with BEV ControlNet~\cite{zhang2023adding}).  We additionally include a \emph{black-box transferability case study} against VAD~\cite{jiang2023vad}, an end-to-end driving model that directly outputs planned trajectories from camera inputs; adversarial samples are optimized solely with MapTR gradients and evaluated on VAD with no access to its weights (\S\ref{sec:transfer}).

\vspace{2pt}\noindent\textbf{Baselines.} We compare against two white-box baselines: (1)~Pixel PGD~\cite{madry2018towards}: standard $L_\infty$ projected gradient descent on input pixels ($\epsilon{=}0.1$); and (2)~AdvPatch~\cite{lou2025asymmetry}: the state-of-the-art adversarial patch attack on HD map construction (CCS~'25), which optimizes a 60$\times$120\,px learnable patches placed roadside. We include two independent patches presented in both the front and back cameras.

\vspace{2pt}\noindent\textbf{Attack Configuration.} \sysname uses partial-diffusion strength $s=0.3$ (6 of 20 denoising steps), 30 PGD iterations, $\epsilon{=}0.5$ in latent space, $\alpha{=}0.05$. CLIP direction weight $\lambda_{\text{CLIP}} \in \{0.3, 0.5\}$, negative anchor weight $\lambda_{\text{neg}}=0.5$. Detection threshold is 0.3. All experiments use 2$\times$ NVIDIA A100-40GB GPUs. Additional implementation details are included in Appendix~\ref{app:implementation}.

\subsection{Evaluation Metrics}

We report both perception and planning-level metrics averaged across frames to measure attack effectiveness. 

\vspace{2pt}\noindent\textbf{Perception Metrics.} We report changes in detections of road boundary element---reduction for boundary removal and increase for boundary injection, in both percentage (\%) and numbers ($\Delta$det).

\vspace{2pt}\noindent\textbf{Planning-level Metrics}: $(1)$ UPTR (Unsafe Plan Trajectory Rate): fraction of scenarios in which the A* path cost under the adversarial BEV prediction \emph{strictly exceeds} the cost under the clean BEV prediction (using the same planner configuration---identical start/goal, grid resolution, and per-element costs described in Appendix~\ref{app:implementation}). 
$(2)$ ORR (Off-Road Rate): fraction of scenarios in which the planned path crosses a boundary polyline from the clean MapTR detection set.
$(3)$ FSR (False-Stop Rate): fraction of scenarios in which the planned trajectory stops $>$5\,m before its goal or deviates $>$3.5\,m laterally from the clean path.

\subsection{Attack Effectiveness}
\label{sec:effectiveness}

\input{tabs/tab_boundary_removal_new}

\subsubsection{Boundary Removal.} Table~\ref{tab:main} presents boundary removal results. \sysname achieves 57.7\% detection drop (8.14$\to$3.44 detections per scene), substantially outperforming AdvPatch (19.2\%) and approaching pixel PGD (72.0\%). All three attacks corrupt the A* planner's trajectory in $\geq$95\% of scenarios (UPTR), but \sysname and PGD suppress far more detections. The key distinction is in defense resilience: PGD's 72\% drop is neutralized by median filtering (80.7\% recovery), while \sysname's 57.7\% drop survives the same defense (35.7\% recovery)---see \S\ref{sec:defense}. 
At the planner level, \sysname approaches the strongest PGD baseline, achieving up to 52\% ORR and 33\% FSR.

\subsubsection{Boundary Injection}
\label{sec:injection}

\input{tabs/tab_boundary_injection_new}

Table~\ref{tab:injection} reveals a striking asymmetry across attack methods for boundary injection. Only \sysname successfully injects fictitious detections, increasing boundary detections by +1.88 per scene on average (9.01$\to$10.89). Pixel PGD, despite its strong boundary removal performance, actually \emph{reduces} total detections ($-1.82$/scene) when targeting boundary injection---its additive pixel noise disrupts the mapping model's existing detections rather than coherently creating new boundary-like features. AdvPatch has negligible effect ($-0.06$/scene). 
At the planner level, \sysname is the strongest attack on every safety metric: ORR $+15$\% (vs.\ pixel PGD $11$\%, AdvPatch $3$\%) and FSR 33\% (vs.\ pixel PGD 25\%, AdvPatch 5\%). 
The injection-side leadership is the safety-relevant claim, since unwarranted emergency stops are immediate collision hazards: pixel PGD's image-space noise reliably suppresses detections but cannot inject coherent boundary-like features; AdvPatch's localized patch produces neither effect strongly; \sysname produces both.

This result highlights a fundamental advantage of semantic-level attacks: the diffusion model can coherently synthesize boundary-like visual features across the scene, producing image content that the mapping model classifies as genuine road boundaries, while pixel-level perturbations lack such structural coherence.

\subsection{Ablation Study}
\label{sec:epsilon}

\input{tabs/tab_perturb_sens}

\subsubsection{Perturbation Budget Sensitivity.} Table~\ref{tab:epsilon} shows detection change across latent-space perturbation budgets for both attack goals on a smaller 3-scene experiments. 
For boundary removal, suppression grows with $\epsilon$ up to 1.0 (77\% drop at the largest budget). Even at $\epsilon{=}0.08$, \sysname suppresses 45.5\% of detections with PSNR above 39\,dB. CLIP shadow guidance holds effectiveness to within $\pm 5$ percentage points of the unconstrained baseline, confirming that semantic steering is effectiveness-neutral for boundary removal. For boundary injection, the detection change rate grows from $+12.5\%$ at $\epsilon{=}0.08$ to $45.8\%$ at $\epsilon{=}0.5$ and then plateaus. 
Notably, CLIP shadow steering \emph{increases} injection effectiveness at $\epsilon \geq 0.5$ (58.3 vs.\ 45.8\% at $\epsilon{=}0.5$), with negligible PSNR cost. This suggests that shadow-like perturbations on straight roads provide linear structural cues that actively help the mapping model hallucinate boundary-like features---CLIP guidance is effectiveness-\emph{positive} for injection, not merely effectiveness-neutral as in the removal case. This finding reinforces that \sysname's semantic attacks are materially different in mechanism from additive noise: the attack benefits from physically plausible scene content, not despite it.

\input{tabs/tab_clip_ablation}

\subsubsection{CLIP Direction Ablation.}\label{sec:clip_ablation} Table~\ref{tab:clip} presents the results ablating the CLIP direction steering. For boundary removal, CLIP direction loss is effectiveness-neutral---shadow matches the unconstrained baseline (57.7\%) and wet-road is within $\sim\!4$ points (53.8\%). CLIP guidance redirects the perturbation toward a specific semantic direction without a measurable effectiveness cost. For boundary injection, CLIP guidance is effectiveness-\emph{positive}: shadow steering increases per-scene $\Delta$ detections from $+1.88$ (no CLIP) to $+2.10$ (+23.4\% of clean detections), and wet-road further to $+2.15$ (+23.9\%). Changes in shadow and wet textures on road provide visual cue that pushes the mapping model to instantiate boundary queries. Semantic steering thus simultaneously improves realism (\S\ref{sec:realism}) and measurably boosts injection effectiveness.

\subsection{Defense Evasion Analysis}
\label{sec:defense}

\input{tabs/tab_defense_evasion}

Table~\ref{tab:defense} presents attack effectiveness under various defenses. 
For boundary removal, \sysname's suppressed count of $3.44$ is stubbornly hard to lift. JPEG-75 recovers it only to $4.35$, Med-3 to $5.12$, and DiffPure to $4.97$---all still far below the clean $8.14$. The same three defenses restore pixel PGD from $2.28$ to $5.45$, $7.01$, and $6.99$ respectively, essentially back to clean, and they push AdvPatch's already-weak $6.58$ to $7.61$--$7.93$. Across the three standard defenses, the restoration gap for \sysname ($0.91$--$1.53$ detections lifted, out of $4.70$ suppressed) is roughly $3$--$4\times$ smaller than for PGD ($3.17$--$4.73$ detections lifted, out of $5.86$ suppressed). 
For boundary injection, only \sysname meaningfully injects while PGD and AdvPatch produce Adv~$\le$~Orig, so the baseline rows in the lower block are included for completeness but do not admit a ``recovery'' reading. Focusing on \sysname: its inflated $10.89$ per-scene count is only lightly reduced by JPEG-75 ($0.13$ of $1.88$ injected detections removed) and Med-3 ($0.60$ removed). DiffPure is substantially more effective, reaching $9.42$ ($1.47$ removed, $\sim\!78\%$ of the injection). This cross-direction pattern is itself informative: the same DiffPure pass that fails against \sysname's removal (leaving $4.97$ well below the clean $8.14$) is more effective against \sysname's injection (yet without a full recovery, leaving $24$\% of the injected boundaries intact)---purification partially smooths away the coherent, synthesized boundary-like features that drive false positives, while the latent-space shadow and texture cues that drive false negatives survive. Nevertheless, no single tested defense simultaneously closes both gaps.

\subsection{Perceptual Realism Validation}
\label{sec:realism}

A core claim of \sysname is that the semantic adversarial images are \emph{realistic}, corresponding to plausible environmental conditions. We validate this claim through two complementary experiments: (1)~Demonstrating that identical perception failures occur naturally in real-world driving, and (2)~Using vision-language models (VLMs) as automated oracles to quantify realism of generated samples.

\subsubsection{Natural Mapping Failures.} We first ask how often MapTR fails on \emph{unmodified} nuScenes data, and---more importantly---how often those failures translate into \emph{planner-relevant} off-road behaviour. On 6,008 samples of the nuScenes test split, 5,860 samples ($97.5\%$) have at least one missed boundary or divider; the per-sample mean of \emph{missed boundaries} is 2.60. This number captures \emph{any} detection miss---most are minor polyline truncations that the planner absorbs and does not reflect a planner-relevant failure rate. To obtain a fairer comparison for attack-induced ORR, we selected the top-50 worst samples in each of three failure-mode (miss-dominant, hallucination-dominant, balanced miss/hallucination) and ran the A* planner used in our attack evaluation. The natural ORR on these 150 samples reach 28\% on the worst-mixed and 24\% on the worst-missing---smaller but on a similar order as \sysname induced ORR (Tables~\ref{tab:main}, \ref{tab:injection}). The vulnerability \sysname exploits is therefore not exotic: adversarial perturbations push the perception model \emph{further along the existing failure axis it exhibits on real, unmodified data}.

To extend the natural-failure observation beyond nuScenes itself, we collect real-world driving data using an Insta360 X3 360$^\circ$ camera mounted on a Tesla Model 3, transformed and calibrated into 6-camera nuScenes format (Appendix~\ref{app:insta360}). We observe the same failure modes from the mapping model---missing boundaries, spurious injected detections, and misclassified road elements---under naturally occurring challenging conditions (\eg strong shadows), as shown in Appendix~\ref{app:insta360}. The combination of these two observations (in-distribution natural failure rate of 24--28\% on worst samples; same qualitative failure modes on out-of-domain real footage) makes a strong case that the vulnerabilities \sysname discovers reflect genuine perception fragility, not generator-specific artifacts.

\subsubsection{VLM Realism Evaluation.} To systematically evaluate perceptual realism, we employ two different open-source VLMs: InternVL3-8B~\cite{internvl3} and Gemma-4-E4B~\cite{gemma4}, as independent realism oracles.\looseness=-1 

\vspace{2pt}\noindent\textbf{Setup.} Each judge receives an expert-persona system prompt and a 4-part user prompt (verbatim in Appendix~\ref{app:realism_prompt}) asking for: (1) a scene description, (2) up to three realism indicators, (3) a single-word YES/NO verdict, and (4) a 1--5 confidence score. 
We evaluate the following five categories each with 200 images: (1)~clean nuScenes originals (100 samples for boundary-removal + 100 injection), (2)~challenging real-world frames uniformly strided from our real-world recording (3)-(5)~adversarial images on the same nuScenes samples attacked with \sysname, PGD, and AdvPatch, respectively. Judges see images blind, without category labels. 

\input{tabs/tab_vlm_realism}

\input{figs/fig_realism}

\vspace{2pt}\noindent\textbf{Results.} As shown in Table~\ref{tab:vlm} and Figure~\ref{fig:realism_bar}, \sysname generated semantic adversarial samples pass as realistic 79.9--84.4\% of the time, within 14--17 percentage points of clean nuScenes (96.5--98.5\%) and close to real-world footage (100\%). Pixel PGD is partially visible to both judges---InternVL3 flags 48\% of PGD images as unrealistic and Gemma flags 72\%, citing ``geometric noise overlays,'' ``high-frequency patterned noise,'' and ``speckled patterns inconsistent with natural camera capture.'' AdvPatch is trivially spotted: only 8.5\% (InternVL3) and 0.0\% (Gemma) of patched images pass, with judges consistently identifying the patch as ``a block of random, high-frequency, multi-colored noise'' or ``digital corruption.''

Despite using different model architectures, both judges produce the same category ordering ($\text{AdvPatch} \ll \text{PGD} < \sysname \lesssim \text{clean} \lesssim \text{challenging}$). Cohen's $\kappa$ on per-image majority verdicts is 0.12--0.18 across non-unanimous categories. Cohen's $\kappa$ is known to be pessimistic under highly skewed base rates~\cite{feinstein1990high}: when both judges agree YES on the vast majority of clean images, $\kappa$ penalizes the residual disagreement on borderline cases more than the agreement on easy ones, producing low values even when the per-image decisions are largely consistent. The claim our realism evaluation supports is the \emph{ordinal} ranking---\sysname is judged closer to clean than PGD or AdvPatch is---and that ranking is fully reproduced by both judges independently. The implication is sharper than the raw $\kappa$ suggests: VLM-based realism verification, even with judge ensembling, would not flag \sysname samples as adversarial.

\vspace{2pt}\noindent\textbf{Implication.} We do not claim that \sysname is undetectable by any method. We claim that the same input-preprocessing defenses that largely neutralize $L_p$ attacks (median filtering recovers 80.7\% of pixel PGD's suppressed detections) leave the majority of \sysname's attack effect intact (35.7\% recovery), and that two independent VLM judges agree on the ordinal realism ranking despite low absolute $\kappa$ on heavily skewed base rates. Standard defense assumptions therefore fail on \sysname at two levels: (i)~pixel-level defenses partially close the gap but leave most of \sysname's effect (\S\ref{sec:defense}), and (ii)~VLM-based semantic verification---a natural fallback defense---also rates \sysname's outputs as realistic at rates comparable to clean nuScenes. The adversarial semantic changes \sysname discovers are latent-space neighbors of real nuScenes scenes drawn from a distribution learned on real driving data; qualitative samples with verbatim VLM outputs are provided in Appendix~\ref{app:realism_prompt}.

%% file: tabs/tab_boundary_removal_new.tex
\definecolor{highlight}{RGB}{242,246,250}

\begin{table}[t]
\caption{\textbf{Boundary removal} results. \sysname achieves 57.7\% detection drop and reduces boundary detection counts by 4.7, comparable to pixel PGD's 72.0\% and 5.9, while requiring substantially different defense strategies (\S\ref{sec:defense}).}
\label{tab:main}
\small
\setlength{\tabcolsep}{3pt}
\resizebox{\columnwidth}{!}{%
\begin{tabular}{lcc|ccc}
\toprule
 & \multicolumn{2}{c|}{\textbf{Perception}} & \multicolumn{3}{c}{\textbf{Planner safety}} \\
\cmidrule(lr){2-3} \cmidrule(lr){4-5}
Method & Drop\,(\%)$\uparrow$ & $\Delta$det$\uparrow$ & ORR\,(\%)$\uparrow$ & FSR\,(\%)$\uparrow$ & UPTR\,(\%)$\uparrow$ \\
\midrule
Pixel PGD & \textbf{72.0} & -5.9 & \textbf{55} & \textbf{34} & 95 \\
AdvPatch~\cite{lou2025asymmetry} & 19.2 & -1.6 & 39 & 20 & 95 \\
\midrule
\rowcolor{highlight}
\textbf{\sysname} & 57.7 & -4.7 & 52 & 33 & \textbf{96} \\
\bottomrule
\end{tabular}%
}
\end{table}

%% file: tabs/tab_boundary_injection_new.tex
\definecolor{highlight}{RGB}{242,246,250}

\begin{table}[t]
\caption{\textbf{Boundary injection} results. Only \sysname successfully injects fictitious boundaries (+1.88/scene) and leads on every planner safety metric.} 
\label{tab:injection}
\small
\setlength{\tabcolsep}{3pt}
\resizebox{\columnwidth}{!}{%
\begin{tabular}{lcc|ccc}
\toprule
 & \multicolumn{2}{c|}{\textbf{Perception}} & \multicolumn{3}{c}{\textbf{Planner safety}} \\
\cmidrule(lr){2-3} \cmidrule(lr){4-6}
Method & Inc. (\%)$\uparrow$ & $\Delta$\,det$\uparrow$ & ORR\,(\%)$\uparrow$ & FSR\,(\%)$\uparrow$ & UPTR\,(\%)$\uparrow$ \\
\midrule
Pixel PGD & -20\% & $-1.82$ & 11 & 25 & 92 \\
AdvPatch~\cite{lou2025asymmetry} & -1\% & $-0.06$ & 3 & 5 & 87 \\
\midrule
\rowcolor{highlight}
\textbf{\sysname} & 21\% & \textbf{+1.88} & \textbf{15} & \textbf{33} & \textbf{93} \\
\bottomrule
\end{tabular}%
}
\end{table}

%% file: tabs/tab_perturb_sens.tex
\begin{table}[t]
\caption{\textbf{Perturbation budget sensitivity} for both attack goals on a smaller 3-scene experiments. Detection change (\%) and image quality (PSNR) as a function of latent-space $\epsilon$, with and without CLIP shadow steering.}
\label{tab:epsilon}
\small
\begin{tabular}{lccccc}
\toprule
 & \multicolumn{2}{c}{\textbf{Detection Change (\%)$\uparrow$}} & \multicolumn{2}{c}{\textbf{PSNR (dB)$\uparrow$}} \\
\cmidrule(lr){2-3} \cmidrule(lr){4-5}
$\epsilon$ & Baseline & + CLIP & Baseline & + CLIP \\
\midrule
\multicolumn{5}{l}{\emph{Boundary removal}} \\
0.08 & 45.5 & 45.5 & 40.8 & 39.8 \\
0.3  & 59.1 & 54.5 & 33.4 & 31.4 \\
0.5  & 54.5 & \textbf{59.1} & 30.5 & 27.7 \\
1.0  & 77.3 & 72.7 & 22.8 & 21.7 \\
\midrule
\multicolumn{5}{l}{\emph{Boundary injection}} \\
0.08 & 12.5 & 20.8 & 38.5 & 38.1 \\
0.3  & 29.2 & 25.0 & 29.8 & 29.8 \\
0.5  & 45.8 & \textbf{58.3} & 26.8 & 26.7 \\
1.0  & 41.7 & \textbf{54.2} & 20.5 & 20.6 \\
\bottomrule
\end{tabular}
\end{table}

%% file: tabs/tab_clip_ablation.tex
\begin{table}[t]
\caption{CLIP direction loss ablation. CLIP guidance is effectiveness-neutral for boundary removal and effectiveness-positive for boundary injection.}
\label{tab:clip}
\small
\begin{tabular}{lccc}
\toprule
\textbf{Configuration} & $\mathbf{\lambda}_\text{CLIP}$ & \textbf{Detection Change (\%)$\uparrow$} & \textbf{PSNR (dB)$\uparrow$} \\
\midrule
\multicolumn{4}{l}{\emph{Boundary removal} (suppression \%)} \\
\sysname baseline        & 0   & 57.7\%            & 30.5 \\
+ CLIP shadow            & 0.5 & 57.7\%            & 26.6 \\
+ CLIP wet-road          & 0.5 & 53.8\%            & 26.7 \\
\midrule
\multicolumn{4}{l}{\emph{Boundary injection} ($\Delta$ det.\ per scene / injection \%)} \\
\sysname baseline        & 0   & 20.9\% & 26.8 \\
+ CLIP shadow            & 0.5 & 23.4\% & 26.6 \\
+ CLIP wet-road          & 0.5 & 23.9\% & 26.6 \\
\bottomrule
\end{tabular}
\end{table}

%% file: tabs/tab_defense_evasion.tex
\definecolor{highlight}{RGB}{242,246,250}

\begin{table}[t]
\caption{\textbf{Defense evasion} results. Mean per-scene detection counts under each input-preprocessing defense. \sysname's removed or injected boundaries remain potent after different defenses.}
\label{tab:defense}
\small
\setlength{\tabcolsep}{4pt}
\begin{tabular}{lccccc}
\toprule
\textbf{Method} & \textbf{Orig} & \textbf{Adv} & \textbf{JPEG-75} & \textbf{Med-3} & \textbf{DiffPure} \\
\midrule
\multicolumn{6}{l}{\textit{Boundary removal}~---~Detection counts $\downarrow$} \\
\addlinespace[1pt]
\quad Pixel PGD            & 8.14 & \textbf{2.28} & 5.45 & 7.01 & 6.99 \\
\quad AdvPatch             & 8.14 & 6.58 & 7.61 & 7.93 & 7.89 \\
\rowcolor{highlight}
\quad \sysname             & 8.14 & 3.44 & \textbf{4.35} & \textbf{5.12} & \textbf{4.97} \\
\midrule
\multicolumn{6}{l}{\textit{Boundary injection}~---~Detection counts $\uparrow$} \\
\addlinespace[1pt]
\quad Pixel PGD  & 8.97 & 7.15 & 8.58 & 9.20 & 8.42 \\
\quad AdvPatch   & 8.97 & 8.91 & 8.97 & 8.94 & 8.96 \\
\rowcolor{highlight}
\quad \sysname               & 8.97 & \textbf{10.89} & \textbf{10.76} & \textbf{10.29} & \textbf{9.42} \\
\bottomrule
\end{tabular}
\end{table}

%% file: tabs/tab_vlm_realism.tex
\definecolor{highlight}{RGB}{242,246,250}

\begin{table}[t]
\caption{\textbf{VLM realism judgment}. Each cell reports the realism rate (\% judged YES) with mean confidence (1--5) in parentheses. Two independent VLM judges agree on the category ordering. \sysname is judged realistic at rates notably higher than PGD and AdvPatch while comparable to clean nuScenes by both independent VLM judges.}
\label{tab:vlm}
\small
\begin{tabular}{lcc}
\toprule
Image category & InternVL3-8B & Gemma-4-E4B \\
\midrule
Clean nuScenes & 96.5\% (4.89) & 98.5\% (4.93) \\
Custom Real-world & 100.0\% (5.00) & 100.0\% (5.00) \\
\midrule
Pixel PGD & 52.0\% (4.37) & 28.0\% (4.95) \\
AdvPatch~\cite{lou2025asymmetry} & 8.5\% (4.07) & 0.0\% (5.00) \\
\rowcolor{highlight}
\sysname & \textbf{79.9\%} (4.65) & \textbf{84.4\%} (4.82) \\
\bottomrule
\end{tabular}
\end{table}

%% file: figs/fig_realism.tex
\begin{figure}[t]
  \centering
  \begin{tikzpicture}
    \definecolor{judgeA}{HTML}{1F77B4}
    \definecolor{judgeB}{HTML}{E8833A}
    \begin{axis}[
      width=\columnwidth,
      height=5.4cm,
      ybar=1.2pt,
      bar width=8pt,
      enlarge x limits={abs=0.55},
      ymin=0, ymax=108,
      ytick={0,20,40,60,80,100},
      ylabel={Realism rate (\%)},
      ylabel style={font=\footnotesize, yshift=-4pt},
      xtick={1,2,3,4,5},
      xticklabels={nuScenes, Custom RW, \textsc{Mirage}, PGD, AdvPatch},
      xticklabel style={font=\footnotesize},
      yticklabel style={font=\scriptsize},
      tick align=outside,
      tick pos=left,
      major grid style={dotted, gray!45},
      ymajorgrids,
      axis line style={gray!60},
      legend style={
        at={(0.5,1.03)}, anchor=south,
        legend columns=2,
        draw=gray!50, fill=white,
        font=\footnotesize,
        column sep=6pt,
        /tikz/every even column/.append style={column sep=10pt},
      },
      error bars/y dir=both,
      error bars/y explicit,
      error bars/error bar style={line width=0.45pt, gray!60!black},
      error bars/error mark options={rotate=90, mark size=2pt, line width=0.45pt, gray!60!black},
    ]
      \addplot+[fill=judgeA!80, draw=judgeA!90!black, line width=0.3pt]
        table[x=x, y=pct, y error minus=elo, y error plus=ehi] {
          x   pct    elo    ehi
          1   96.5   2.5    2.5
          2  100.0   0      0
          3   79.9   5.53   5.53
          4   52.0   7.0    6.5
          5    8.5   3.5    4.0
        };
      \addlegendentry{InternVL3-8B}
      \addplot+[fill=judgeB!80, draw=judgeB!80!black, line width=0.3pt]
        table[x=x, y=pct, y error minus=elo, y error plus=ehi] {
          x   pct    elo    ehi
          1   98.5   2.0    1.5
          2  100.0   0      0
          3   84.4   5.02   5.02
          4   28.0   6.0    6.5
          5    0.0   0      0
        };
      \addlegendentry{Gemma-4-E4B}
    \end{axis}
  \end{tikzpicture}
  \caption{\textbf{Per-category realism rates} with 95\% bootstrap confidence intervals. Both VLM judges converge on the same category ordering. \sysname's images are judged realistic at rates comparable to clean nuScenes and challenging real-world footage; PGD and AdvPatch are trivially flagged.}
  \Description{Grouped bar chart comparing two VLM judges' realism rates across five image categories, with bootstrap error bars.}
  \label{fig:realism_bar}
\end{figure}
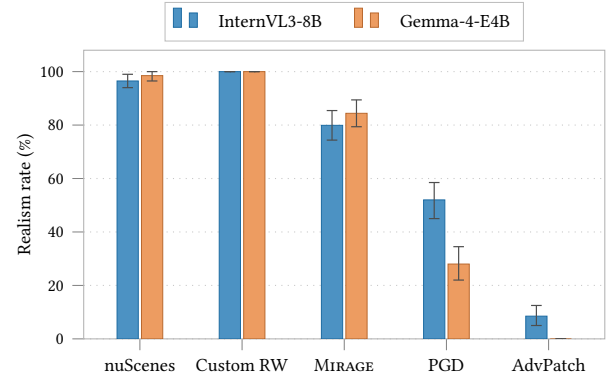

%% file: secs/case_study.tex


\subsection{Transferability to End-to-End Planner}
\label{sec:transfer}


Our formal attack optimizes only through the mapping model. To probe whether semantic vulnerabilities discovered on one perception backbone transfer to an entirely different driving model, we feed the \emph{same} adversarial images produced in~\S\ref{sec:effectiveness} to an end-to-end driving model VAD~\cite{jiang2023vad}. VAD's weights are treated as unknown. 

\input{tabs/tab_vad}

\input{figs/fig_physical_feasibility}

Table~\ref{tab:planner} reports VAD trajectory deviation from its own clean-baseline planning. For boundary removal, \sysname induces the largest average displacement (ADE 0.538\,m vs.\ 0.460\,m for PGD) and maximum deviation (0.888\,m vs.\ 0.764\,m), indicating that its semantic perturbations transfer more effectively than $L_p$ or patch perturbations across perception backbones. 
For boundary injection, absolute deviations across all three attacks are smaller ($\sim$0.44--0.45\,m ADE) and the gap between methods narrows---consistent with the observation that VAD's multi-frame temporal modeling provides partial resilience to single-frame semantic manipulation. Because the attacker here has no knowledge of VAD at all, this result offers preliminary evidence that semantic, distribution-consistent perturbations constitute a black-box threat to end-to-end driving stacks---not a mapping model specific idiosyncrasy. 
Additionally, we note that \sysname's semantic adversarial attack discovery framework enables the attacker to directly search for natural environmental variations leading to unsafe planning, by adapting an end-to-end differentiable driving model in place of the map construction model, which is an important next step we leave to future work.

\subsection{Physical Realizability Case Study}\label{sec:physical_feasibility}

We 
conduct a small 
\emph{physical} case study in which we hand-reproduce \sysname-discovered patterns on a real road and re-collect the data through the same camera rig. We frame this as a feasibility \emph{proof of concept} rather than a rigorous quantitative measurement: the goal is to test whether even a low-fidelity, low-cost physical reproduction of \sysname's optimized patterns is sufficient to perturb mapping model's predictions in the direction \sysname intended.

\subsubsection{Setup.} The case study was conducted on an empty private road section owned and managed by our University, with prior approval from the University Police Department and Facility Management. The data collection uses the same setup as our real-world dataset collection, with an Insta360 X3 360$^\circ$ camera on the roof of a Tesla Model 3 at approximately $1.45$\,m above ground (Figure~\ref{fig:rw_setup}, Appendix~\ref{app:insta360}), matching the nuScenes camera extrinsic. The 360$^\circ$ footage was post-processed into the same six virtual pinhole cameras, downsampled to 424$\times$800 for MapTR inference. We take proper safety precautions and report details in Ethical Considerations.



\subsubsection{Procedure.} (1)~We first recorded a clean pass through the test section and ran MapTR on the resulting frame. 
(2)~Using this frame as the input, we ran \sysname twice---once with the boundary-removal objective and once with the boundary-injection objective---both with CLIP direction guidance steered toward the text prompt ``\emph{patterns on the ground}.'' The optimized adversarial views are shown in Figure~\ref{fig:rwf_views}; relative to the clean inputs, both contain low-contrast surface patterns concentrated on the road in front of the vehicle, in line with the steering prompt. (3)~Two team members then hand-reproduced the dominant ground-pattern features---roughly traced shape, location, and extent---in sidewalk chalk on the asphalt; we did not attempt to match texture, edge sharpness, or color of the optimized patterns. (4)~We re-collected a pass over the chalked road in the same lighting window, processed the new footage through the same six virtual cameras, and ran MapTR again. We label the four resulting BEV predictions in Figure~\ref{fig:rwf_bev}: clean, MapTR run on \sysname's optimized digital views (``digital''), and MapTR run on the chalk-reproduced views (``physical'').

\subsubsection{Observations.}
\begin{figure}[t]
  \centering
  \begin{subfigure}{0.19\columnwidth}
    \centering
    \includegraphics[width=\linewidth]{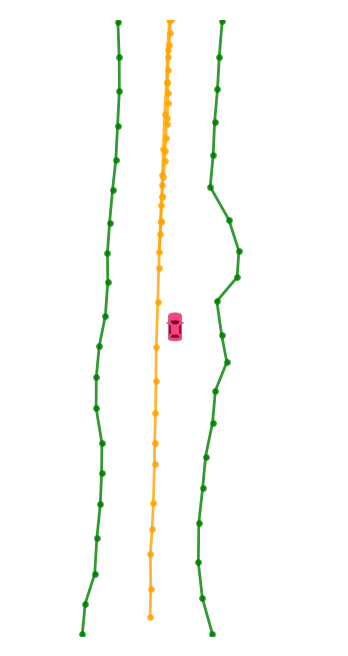}
    \caption{Clean.}
  \end{subfigure}\hfill
  \begin{subfigure}{0.19\columnwidth}
    \centering
    \includegraphics[width=\linewidth]{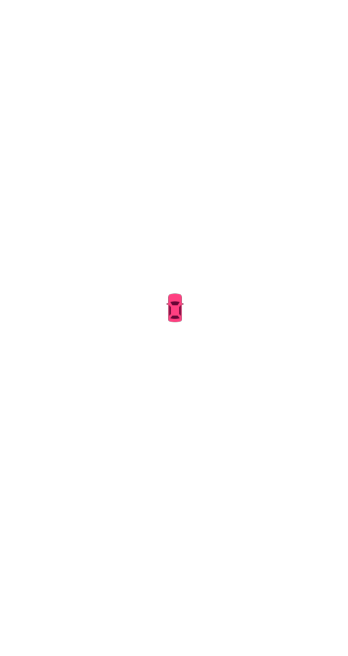}
    \caption{D-rem.}
  \end{subfigure}\hfill
  \begin{subfigure}{0.19\columnwidth}
    \centering
    \includegraphics[width=\linewidth]{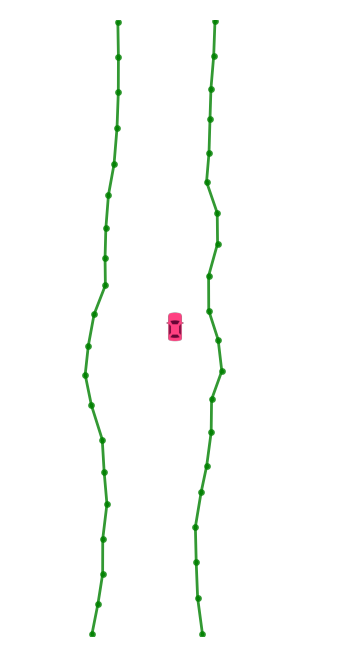}
    \caption{P-rem.}
  \end{subfigure}\hfill
  \begin{subfigure}{0.19\columnwidth}
    \centering
    \includegraphics[width=\linewidth]{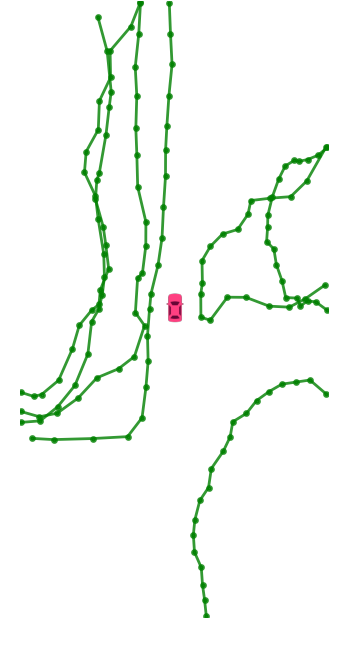}
    \caption{D-inj.}
  \end{subfigure}\hfill
  \begin{subfigure}{0.19\columnwidth}
    \centering
    \includegraphics[width=\linewidth]{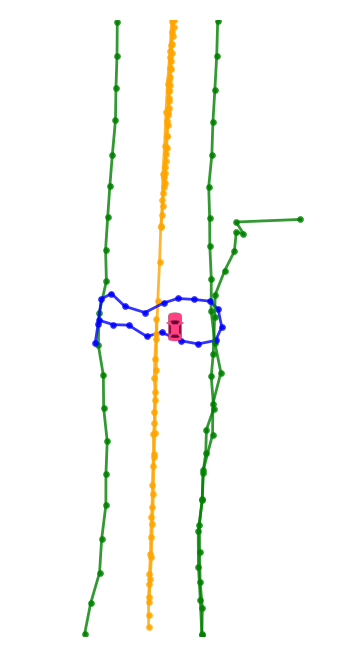}
    \caption{P-inj.}
  \end{subfigure}
  \caption{MapTR BEV outputs for the physical case study. ``D''/``P'' = digital optimized views vs.\ chalk-reproduced physical views; ``rem.''/``inj.'' = removal vs.\ -injection objective.} 
  \Description{Five small bird's-eye-view diagrams showing road element predictions. From left to right: clean (two green boundary lines and one orange central divider), digital removal (no detections), physical removal (boundaries kept, divider gone), digital injection (many phantom boundaries), physical injection (a phantom blue pedestrian crossing in front of the car plus an extra boundary fragment).}
  \label{fig:rwf_bev}
\end{figure}

Comparing the BEV outputs in Figure~\ref{fig:rwf_bev}:
(i)~On the boundary-removal target, the central orange divider that is detected cleanly under normal conditions is fully suppressed in the physical-chalk pass---both green boundaries are still present, but the divider has been removed in the same direction that the digital \sysname pass removes it.
(ii)~On the boundary-injection target, the physical-chalk pass induces a fictitious blue pedestrian crossing polygon around the vehicle, plus a phantom right-side boundary fragment, neither of which exists in the clean prediction. The fictitious boundary injected on the right, while subtle, matches the qualitative direction of the digital \sysname pass on the same scene.
In other words, even with a manual chalk reproduction that captures only a coarse outline of \sysname's optimized texture, the failure mode \sysname predicted is the failure mode the deployed mapping model exhibits. The digital pass remains the upper bound: it removes \emph{all} road elements (Figure~\ref{fig:rwf_bev}b) or hallucinates a denser web of phantom boundaries (Figure~\ref{fig:rwf_bev}d), whereas \textit{the physical pass produces a smaller but qualitatively congruent semantic perturbation}.\looseness=-1

\subsubsection{Faithfulness and Tolerance for Replication Error.}
The chalk-reproduced patterns are visibly coarser than \sysname's optimized output: hand drawing cannot match the diffusion model's fine-grained low-contrast shading, edge softness, or asphalt-color blending. 
Time-of-day also differed between the optimization input and the reproduction pass. Despite this, the directional effect on mapping results is preserved. We read this as evidence that the underlying mapping vulnerability to semantic adversarial conditions is \textit{robust to substantial implementation error} in the physical pattern: the attacker does not need to faithfully reconstruct a pixel-perfect rendering of \sysname's output, only its dominant structural cues. This is consistent with the broader finding in \S\ref{sec:realism} that \sysname's perturbations are not high-frequency artifacts but low-frequency semantic cues, which by construction tolerate pixel-level defenses.

%% file: tabs/tab_vad.tex
\begin{table}[t]
\caption{\textbf{VAD transferability}. Adversarial samples crafted on MapTR are fed zero-shot to VAD (no access to weights or gradients). \sysname's semantic perturbations transfer more effectively than $L_p$ or patch baselines for boundary removal.}
\label{tab:planner}
\small
\begin{tabular}{lccc}
\toprule
 & \textbf{ADE (m)} & \textbf{FDE (m)} & \textbf{MaxDev (m)}  \\
\midrule
\multicolumn{4}{l}{\emph{Boundary removal}} \\
\quad Pixel PGD & 0.460 & 0.746 & 0.764 \\
\quad AdvPatch  & 0.464 & 0.757 & 0.777 \\
\quad \sysname  & \textbf{0.538} & \textbf{0.871} & \textbf{0.888} \\
\midrule
\multicolumn{4}{l}{\emph{Boundary injection}} \\
\quad Pixel PGD & 0.435 & 0.688 & 0.723  \\
\quad AdvPatch  & 0.442 & 0.695 & 0.738 \\
\quad \sysname  & \textbf{0.454} & 0.694 & 0.714 \\
\bottomrule
\end{tabular}
\end{table}

%% file: figs/fig_physical_feasibility.tex
\begin{figure*}[t]
\centering
\begin{tabular}[t]{c|c|c}
\begin{subfigure}[l]{0.5\textwidth}
    \includegraphics[width=\linewidth]{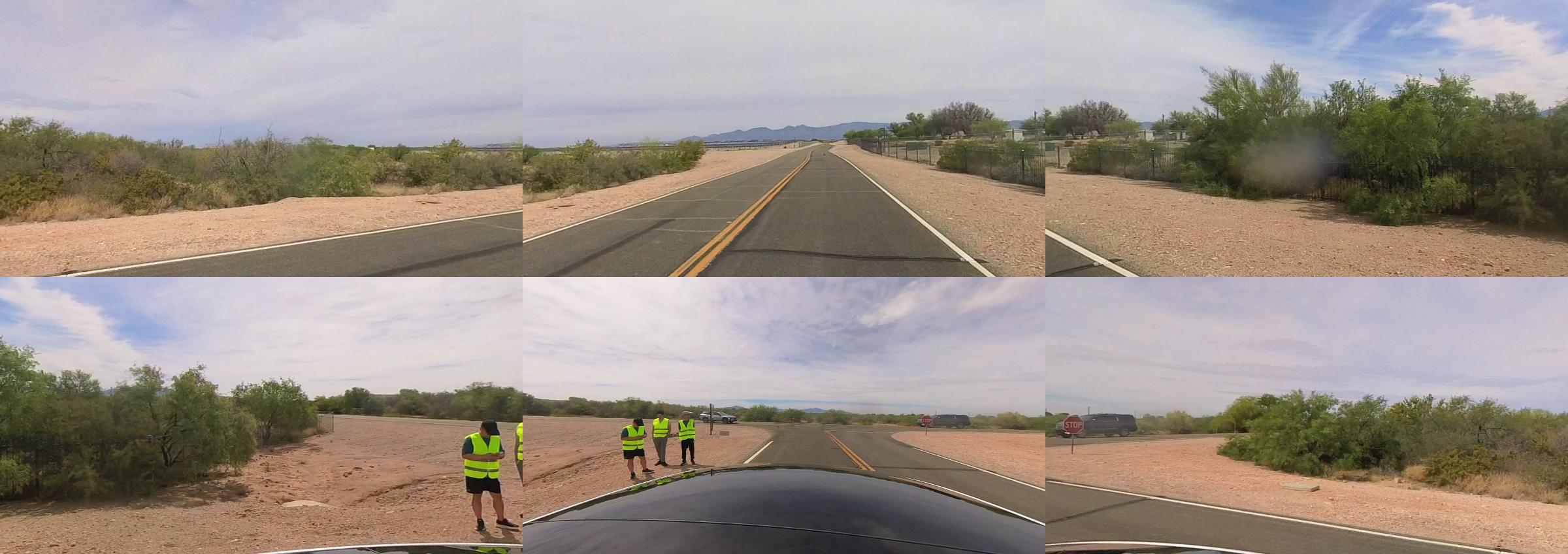}
    \caption{Original views.}
\end{subfigure}
\smallskip
     &  
        \begin{tabular}{c}
        \smallskip
        \begin{subfigure}[t]{0.2\textwidth}
            \includegraphics[width=1.0\linewidth]{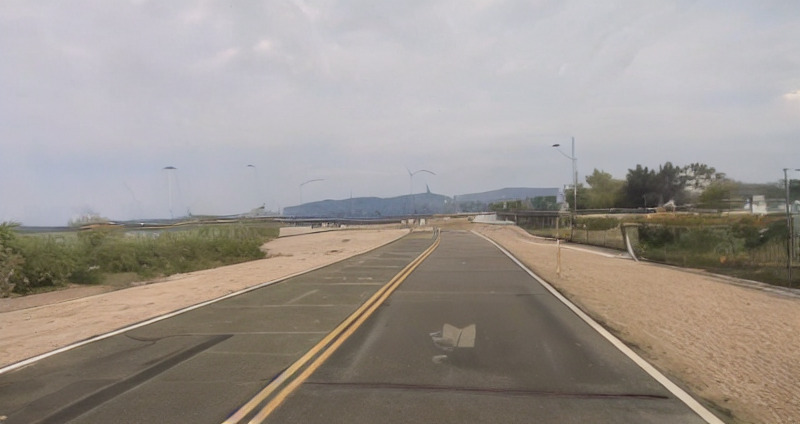}
            \caption{\sysname, removal.}
        \end{subfigure}\\
        \begin{subfigure}[t]{0.2\textwidth}
            \includegraphics[width=1.0\linewidth]{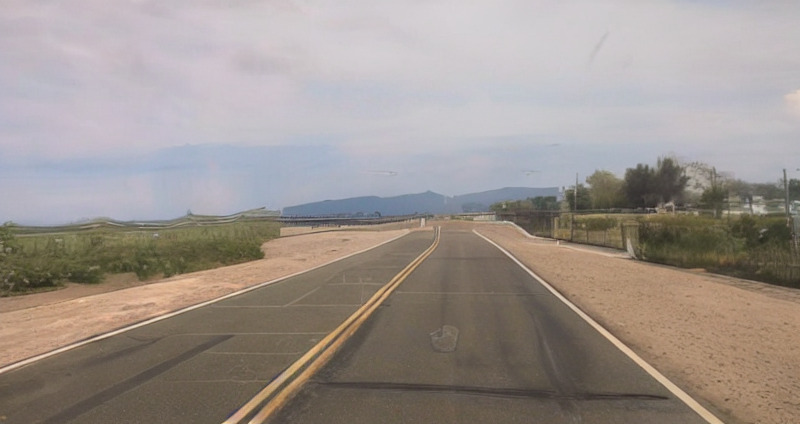}
            \caption{\sysname, injection.}
        \end{subfigure}\\
        \end{tabular}
    &
        \begin{tabular}{c}
        \smallskip
        \begin{subfigure}[t]{0.2\textwidth}
            \includegraphics[width=1.0\linewidth]{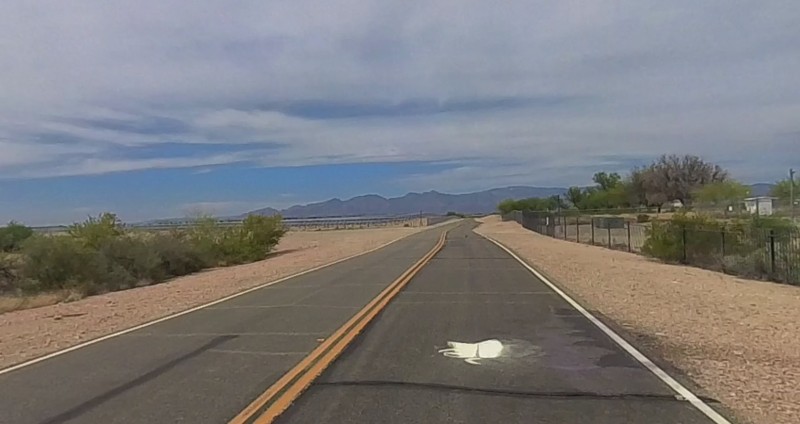}
            \caption{Reproduced, removal.}
        \end{subfigure}\\
        \begin{subfigure}[t]{0.2\textwidth}
            \includegraphics[width=1.0\linewidth]{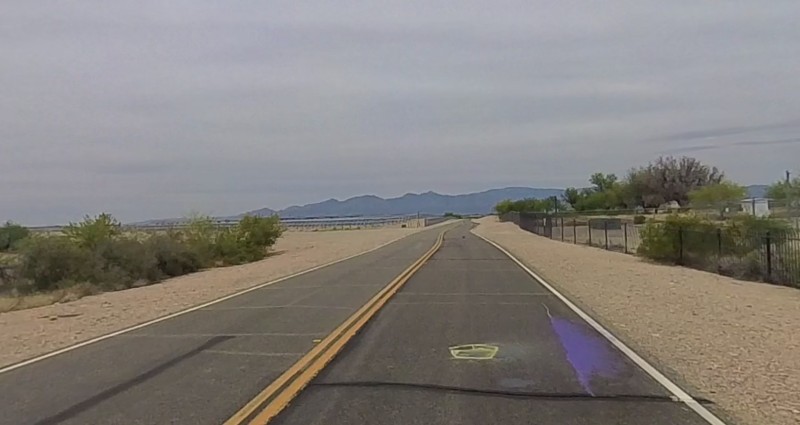}
            \caption{Reproduced, injection}
        \end{subfigure}\\
        \end{tabular}
\end{tabular}
\caption{\textbf{Camera evidence for the physical case study.} (a) Original scenes captured for experiments. (b)-(c) \sysname's optimized digital adversarial views for the boundary-removal and boundary-injection objectives, both with CLIP direction guidance ``\emph{patterns on the ground}.'' (d)-(e) Corresponding views after we hand-reproduce the dominant ground patterns in chalk on the asphalt and re-collect a pass through the same camera rig. The chalk reproduction captures only a coarse outline of the optimized pattern's shape and location, with no attempt to match texture, edge softness, or color. We show front camera views here for clarity--the surround-view images are included in Figure~\ref{fig:rwf_views_surround} in Appendix.}
  \Description{Two rows of six-camera surround driving views: two adversarial views generated by Mirage (boundary removal and injection) and two physical re-captures of the road after chalk markings were applied.}
\label{fig:rwf_views}
\end{figure*}

%% file: secs/discussion.tex
\section{Discussion}
\label{sec:discussion}

\vspace{2pt}\noindent\textbf{Comparison with AdvPatch.} AdvPatch~\cite{lou2025asymmetry} demonstrates a physically realized attack via roadside patches but is limited to localized influence: it suppresses only 19.2\% of detections versus \sysname's 57.7\% on 100 scenes (Table~\ref{tab:main}), and is neutralized by median filtering (86.5\% recovery, Table~\ref{tab:defense}). \sysname discovers a different vulnerability class where road-surface environmental conditions whose influence is distributed across the scene rather than confined to a roadside patch. The two approaches reveal complementary facets of perception fragility, but \sysname's semantic perturbations are inherently harder to defend against because they lack the  high-frequency signature that makes attacks easily detectable.

\vspace{2pt}\noindent\textbf{Implications for Defense Design.} Our defense evaluation reveals a \emph{categorical} gap: every tested defense family operates on the assumption that adversarial perturbations introduce detectable statistical anomalies (high-frequency noise, spatial outliers, distribution shifts). \sysname violates this assumption entirely---its perturbations are discovered from legitimate environmental variation, drawn from the same real-world distribution the diffusion model was trained on. 
Effective defenses require fundamentally different approaches: semantic anomaly detection (\eg verifying shadow consistency with sun position and time of day), temporal consistency enforcement across frames, multi-sensor cross-validation against LiDAR or radar (unaffected by visual semantic changes), or adversarial environmental augmentation during training.

\vspace{2pt}\noindent\textbf{Limitations and Future work.} Our formal evaluation instantiates the attack pipeline on a single generator--mapper pair (MagicDrive~\cite{gao2023magicdrive} as the scene generator and MapTR~\cite{liao2023maptr} as the victim mapping model). Sweeping over additional mapping architectures such as VectorMapNet~\cite{liu2023vectormapnet} is a natural extension we leave to future work---while the \sysname paradigm applies to any conditional diffusion generator and differentiable mapping head.

Our proof-of-concept physical realizaibility case study uses hand-reproduced patterns in chalk, matching the dominant ground patterns from \sysname's optimized output, which is sufficient to steer mapping results (removing/injecting detections) in the same directions the attacker intended. 
Chalk is a deliberately conservative choice: it is readily removable and visible to safety personnel, but it is therefore also more conspicuous to a human driver than the truly low-contrast textures \sysname optimizes. A more capable adversary with advanced manufacturing access would likely produce closer reproductions of the optimized patterns at the cost of more setup. We leave a controlled multi-scene physical evaluation, including a comparison of reproduction media, to future work; the present case study is included as one piece of feasibility evidence that \sysname's discovered vulnerabilities survive the digital-to-physical gap rather than as a definitive measurement of attack rate in the wild. 

%% file: secs/conclusion.tex
\vspace{-1em}\section{Conclusion}
%
We present \sysname, a framework for systematic discovery of semantic vulnerabilities in camera-based HD map perception. \sysname inverts driving images into a diffusion model's latent manifold and searches for nearby latents that degrade perception, yielding physically plausible environmental conditions that suppress or corrupt map element detection. On nuScenes, \sysname suppresses 57.7\% of boundary detections and is the only evaluated method that reliably injects fictitious boundaries (+1.88/scene), leading on every safety-critical planner metric. Meanwhile, standard input preprocessing (JPEG, median, DiffPure) largely neutralizes pixel PGD but leaves most of \sysname's effect intact. Two independent VLM judges rate \sysname's adversarial images as realistic 80--84\% of the time---on par with clean nuScenes (97--99\%) and far above PGD (28--52\%) and AdvPatch (0--9\%)---and the same failure modes appear spontaneously in unmodified nuScenes and our real-world footage. A chalk-reproduced physical case study further shows the discovered failure modes survive the digital-to-physical gap. Our work suggests that autonomous-driving security must address semantic-level robustness in addition to the established $L_p$ threats.

%% file: secs/ethics.tex
\section{Ethical Considerations}
\label{sec:ethics}

\sysname identifies semantic conditions that degrade a safety-critical perception component of production-style ADS stacks. Since the work is fundamentally a vulnerability analysis on a deployed class of systems, we discuss the risk--benefit tradeoff and the steps we took to minimize harm beyond institutional compliance.

\paragraph{Stakeholders and Harms.} The impacted parties are: (i)~drivers and pedestrians sharing roads with camera-only HD-mapping stacks; (ii)~AV vendors and the robust-perception research community; (iii)~third-party road users incidentally captured in our footage; (iv)~the research team; and (v)~society at large. The principal disclosure concern is adversary uplift---an attacker could use \sysname or its qualitative cues (ground-pattern textures, shadow-like configurations) to deliberately mislead a fielded mapping model.

\paragraph{Mitigations and Benefits (Beneficence, Justice).} \sysname does not introduce a novel mutation; it surfaces a sensitivity that already manifests in unmodified data. The same failure modes appear spontaneously in nuScenes and real-world footage we collected (Appendix~\ref{app:insta360}). The vulnerability is therefore being encountered today by drivers and AV operators whether or not it is named, and the artifact \sysname produces---a characterization of environmental conditions that reliably degrade perception---is the same artifact defenders need in order to harden mapping models, design semantic-aware monitors, or constrain operational design domains. Beyond external defenders, \sysname is directly usable by AV developers themselves as an internal red-team and debugging tool: by running it against their own perception stack, developers can systematically surface failure-inducing scenes, investigate the underlying model behavior, prioritize data collection or augmentation in the affected semantic slices, and use the discovered samples for adversarial training, regression testing, and operational design domain refinement. The established $L_p$ threat model is, as our defense evaluation shows, categorically too narrow to drive the defenses this paper motivates. We judge the long-run benefit to defenders and AV developers to outweigh the marginal adversary uplift.

\paragraph{Disclosure.} Our victim model (MapTR~\cite{liao2023maptr}) and scene generator (MagicDrive~\cite{gao2023magicdrive}) are open-source academic releases evaluated on the public nuScenes benchmark; no single vendor owns a coordinated-disclosure clock. 
We release attack code and the realism-study assets to enable reproducibility and defense research, without any artifact whose primary utility is operational misuse against a fielded vehicle.

\paragraph{Physical Experimentation (Respect for Law and Public Interest).} The on-road study (\S\ref{sec:physical_feasibility}) used a private University-owned road approved by the University Police Department and Facility Management. The segment was closed to all third-party traffic; on-foot personnel wore high-visibility vests; traffic cones delineated the test area (Figure~\ref{fig:rw_setup}). The only road modification was water-soluble sidewalk chalk, swept off after each pass; we deliberately chose chalk over more durable or higher-fidelity media (paint, asphalt sealant) because chalk is removable on a timescale of minutes and is more conspicuous to a human driver than the textures \sysname optimizes. No member of the public, no public road, and no public-road traffic was exposed to any modification.

\paragraph{Data Collection.} We collect no human-subjects data: real-world footage was recorded by the authors driving the data-collection vehicle on public roads under normal traffic, with no audio capture. Faces and license plates of incidentally captured third parties are blurred in any released frame, and no PII is associated with the released dataset. The VLM realism study uses open-weight models (InternVL3-8B, Gemma-4-E4B) on the images themselves with no human raters or crowdworker labor. 
 
\paragraph{Decision.} Both \emph{Beneficence} (defender benefit on an already-occurring failure mode outweighs marginal adversary uplift) and \emph{Respect for Persons} (no human-subjects data, blurred third-party PII) point to the same conclusion: conducting and publishing this work is the more ethical path than withholding it.

%% file: secs/appendix.tex
\begin{figure}[t]
    \centering
    \begin{subfigure}[!htp]{0.98\columnwidth}
        \includegraphics[width=\linewidth]{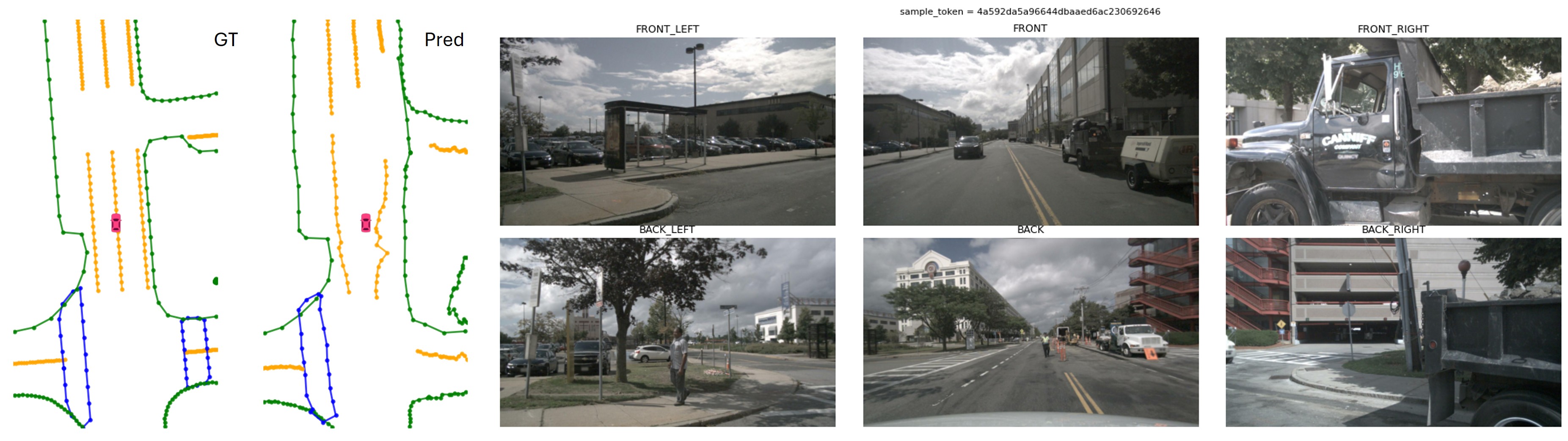}
        \caption{Missing detection of the center boundary.}
    \label{fig:nuscenes_missed}
    \end{subfigure}
    \begin{subfigure}[t]{0.98\linewidth}
        \includegraphics[width=\linewidth]{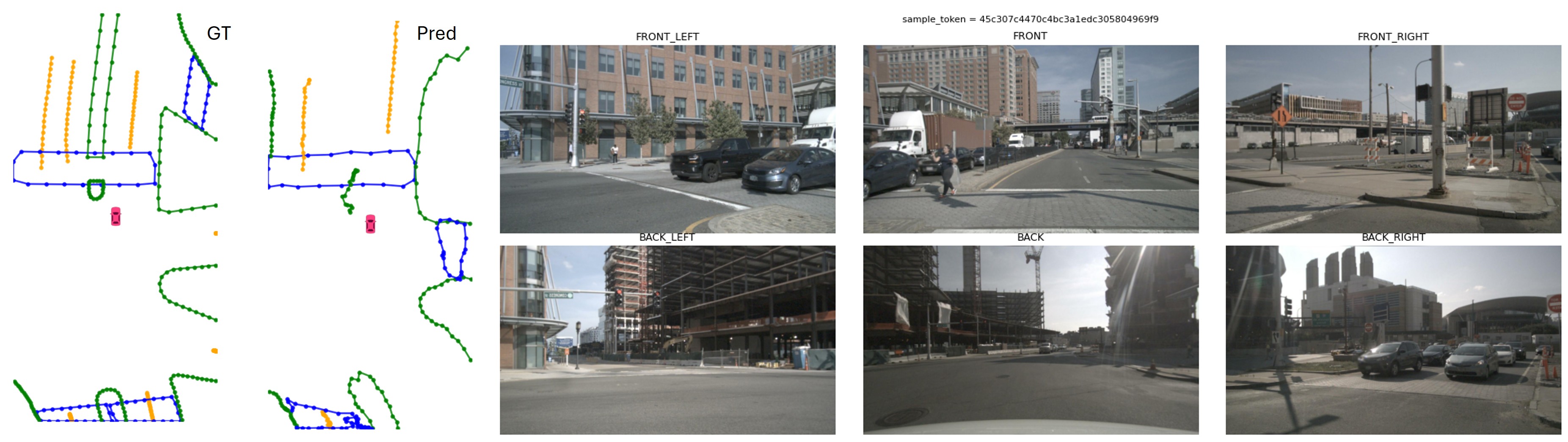}
    \caption{Missing detection of the center boundary, while hallucinating an non-existing boundary  pattern.}
    \label{fig:nuscenes_missed_hallucinated}
    \end{subfigure}
    \Description{nuScenes images showing MapTR's incorrect inference results on natural examples.}
    \caption{Examples of incorrect mapping results on original nuScenes images.}
    \label{fig:nuscenes_error}
    \vspace{-5pt}
\end{figure}

\begin{figure}[t]
    \centering
    \begin{subfigure}[!htp]{0.98\columnwidth}
        \includegraphics[width=\linewidth]{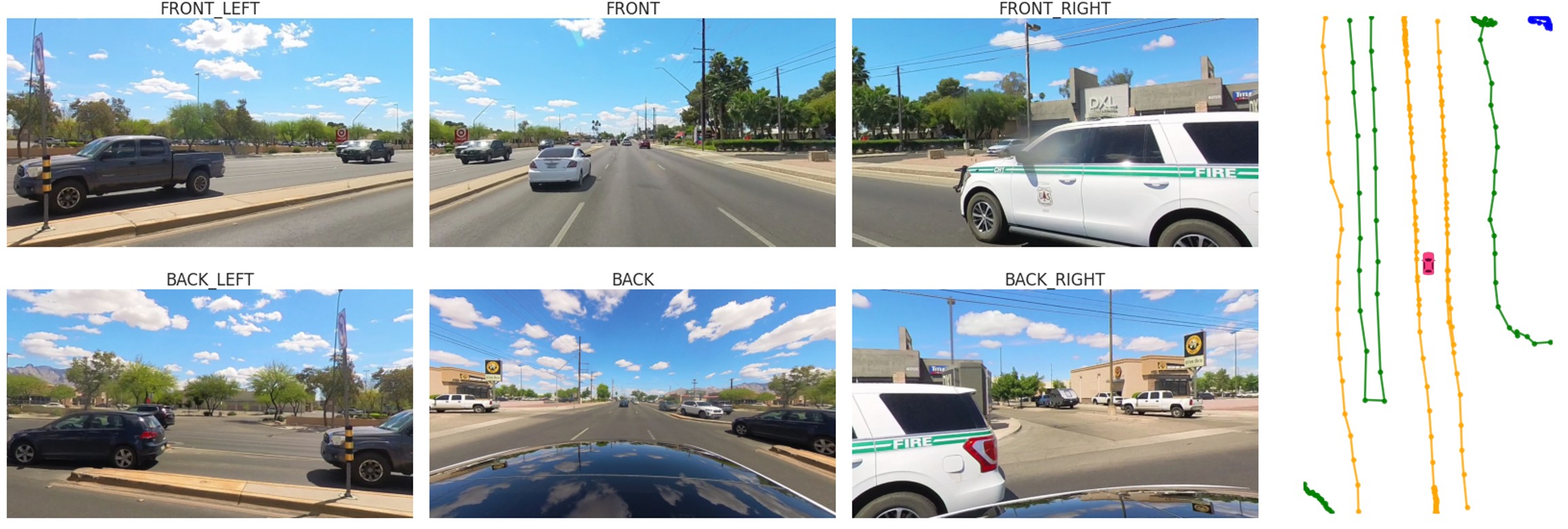}
        \caption{Normal detection of road elements.}
    \label{fig:rw_normal}
    \end{subfigure}
    \begin{subfigure}[t]{0.98\columnwidth}
        \includegraphics[width=\linewidth]{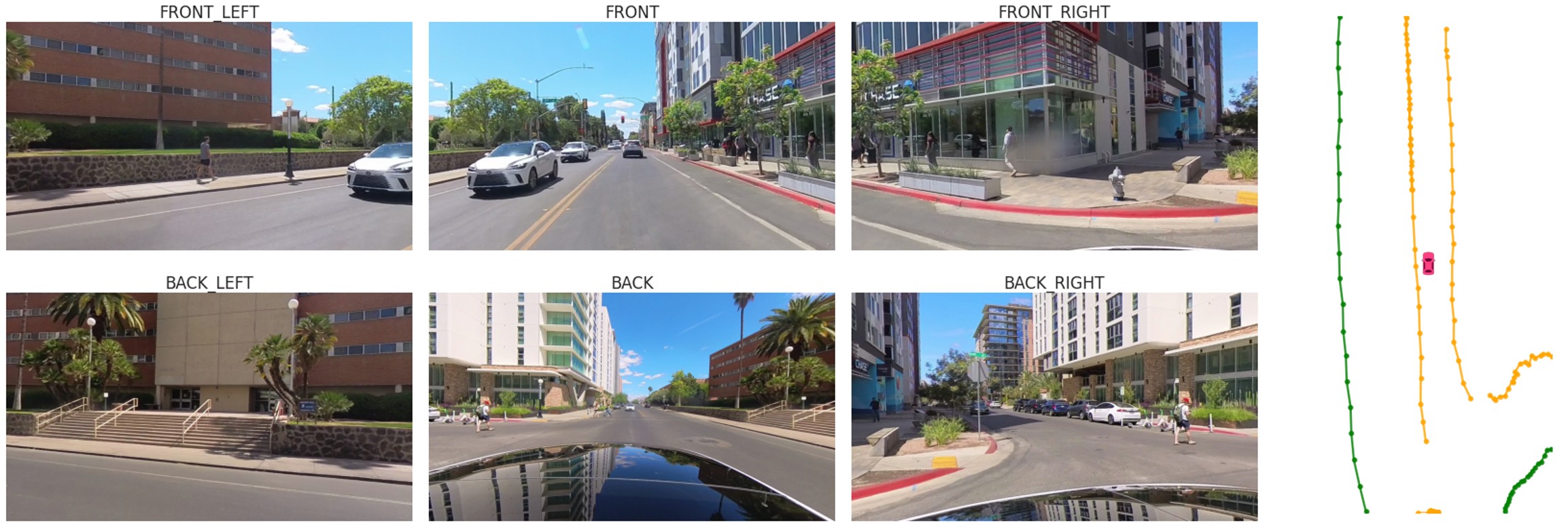}
    \caption{Missing detection of the right boundary.}
    \label{fig:rw_miss}
    \end{subfigure}
    \begin{subfigure}[t]{0.98\columnwidth}
        \includegraphics[width=\linewidth]{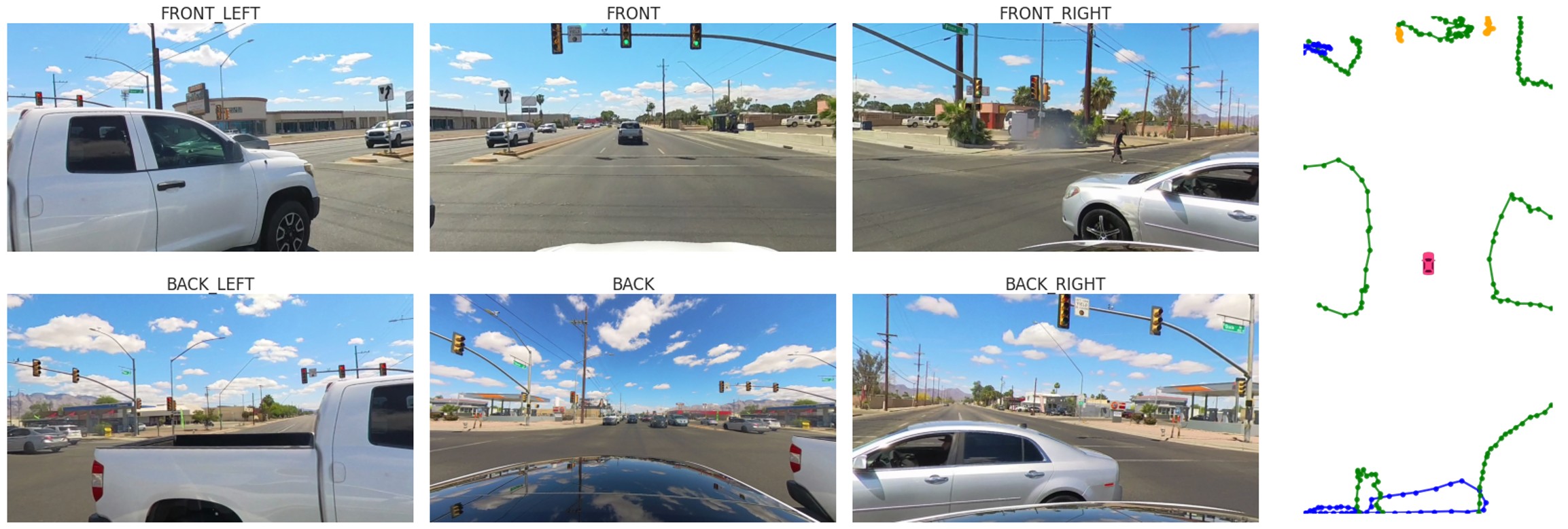}
    \caption{Hallucinated road boundary detection.}
    \label{fig:rw_hallucinate}
    \end{subfigure}
    \Description{Real-world images showing MapTR's incorrect inference results on natural examples.}
    \caption{Examples of incorrect mapping results on real-world footage we collected.}
    \label{fig:rw_error}
    \vspace{-5pt}
\end{figure}

\section{Implementation Details}
\label{app:implementation}

\paragraph{Multi-GPU Setup.} MagicDrive (UNet + text encoder) runs on GPU 0; ControlNet, VAE decoder, and MapTR run on GPU 1. CLIP ViT-L/14 vision encoder also runs on GPU 1 ($\sim$1.2\,GB). Gradient checkpointing is used for per-image VAE decoding and UNet/ControlNet blocks. The split classifier-free guidance strategy runs the unconditional pass without gradients, halving per-step VRAM.

\paragraph{A* Planner.} BEV cost map at 0.25\,m resolution, 120$\times$240 cells covering [$-15$, 15]$\times$[$-30$, 30]\,m. Dividers: cost 50, boundaries: cost 200, free space: cost 1. 8-connected grid with curvature penalty (2.0) and reverse penalty (5.0). ORR is computed by checking whether any point on the planned path falls outside the ground-truth road polygon; FSR checks whether the path terminates early (stop) or deviates laterally by $>$3.5\,m from the original trajectory (deviate).

\paragraph{VAD Transferability Case Study.} Adversarial samples optimized solely with MapTR gradients (no access to VAD weights or gradients at any point in the pipeline) are fed zero-shot to VAD, and we extract its planned 3-second trajectory. ADE/FDE/MaxDev are computed relative to VAD's clean-baseline trajectory; ORR and FSR are computed against the same ground-truth road boundaries used for A* evaluation. 

\paragraph{AdvPatch Baseline.} We reimplement the white-box AdvPatch attack from~\cite{lou2025asymmetry} following their published methodology: a learnable rectangular patch (60$\times$120 pixels) is optimized via PGD and placed on a non-road surface visible from 1--2 cameras. The patch is constrained to not overlap road pixels.

\paragraph{Compute Resources.} All experiments run on 2$\times$ NVIDIA A100-SXM4-40GB GPUs. Per-sample attack time: $\sim$8\,min for 30 PGD steps. Total compute for all reported experiments: approximately 50 GPU-hours.

\section{Real-World Data Collection}
\label{app:insta360}

\begin{figure}[!htb]
  \centering
    \begin{subfigure}[t]{0.23\textwidth}
        \includegraphics[width=\linewidth]{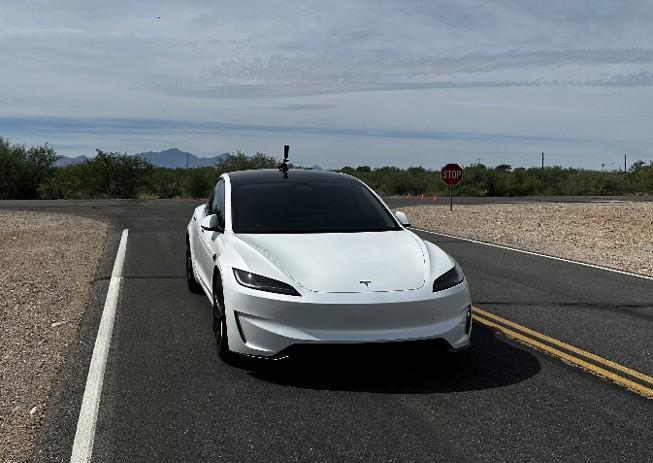}
        \caption{Vehicle and camera rig.}
    \end{subfigure}
    \begin{subfigure}[t]{0.23\textwidth}
        \includegraphics[width=\linewidth]{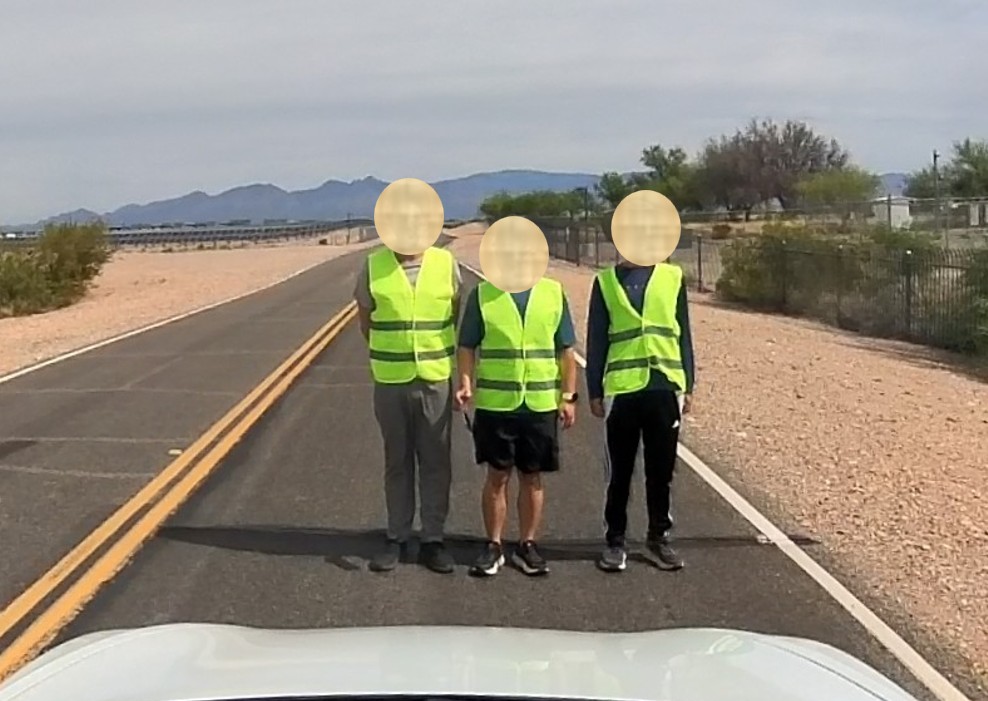}
        \caption{Experiment personnels.}
    \end{subfigure}
    \label{fig:rwf_setup_a}
  \caption{Insta360 X3 mounted on the roof-top of a Tesla Model 3 at ${\sim}1.45$\,m, calibrated to the nuScenes camera height. On-foot personnel in high-visibility vests; the road segment is closed to third-party traffic}
  \Description{white Tesla Model 3 with a 360-degree camera mounted on the roof, parked on a closed two-lane road.}
  \label{fig:rw_setup}
\end{figure}

To validate that \sysname's discovered vulnerabilities correspond to real-world perception failures, we collect driving data using an Insta360 X3 360$^\circ$ camera mounted on the roof of a Tesla Model 3, as shown in Figure~\ref{fig:rw_setup}. The 360$^\circ$ footage is split into 6 virtual pinhole cameras matching the nuScenes camera configuration (front, front-left, front-right, back, back-left, back-right) using known intrinsic and extrinsic parameters. We calibrate the virtual cameras to approximate the nuScenes field-of-view and resolution (900$\times$1600, downsampled to 424$\times$800 for MapTR inference).

We record approximately 20 minutes of urban and suburban driving across varying conditions: sunny, overcast, strong shadows (low sun angle), wet road surfaces, and mixed lighting (tunnels, overpasses). From this footage, we extract keyframes at 2\,Hz and run MapTR-Tiny inference on each. We manually identify ``challenging'' frames where MapTR produces clearly incorrect predictions---missing boundaries near strong shadows, spurious detections on unusual surfaces, and misclassified road elements---and verify these against the actual road layout.

\begin{figure}[t]
  \centering
  \begin{subfigure}{0.48\textwidth}
    \centering
    \includegraphics[width=\linewidth]{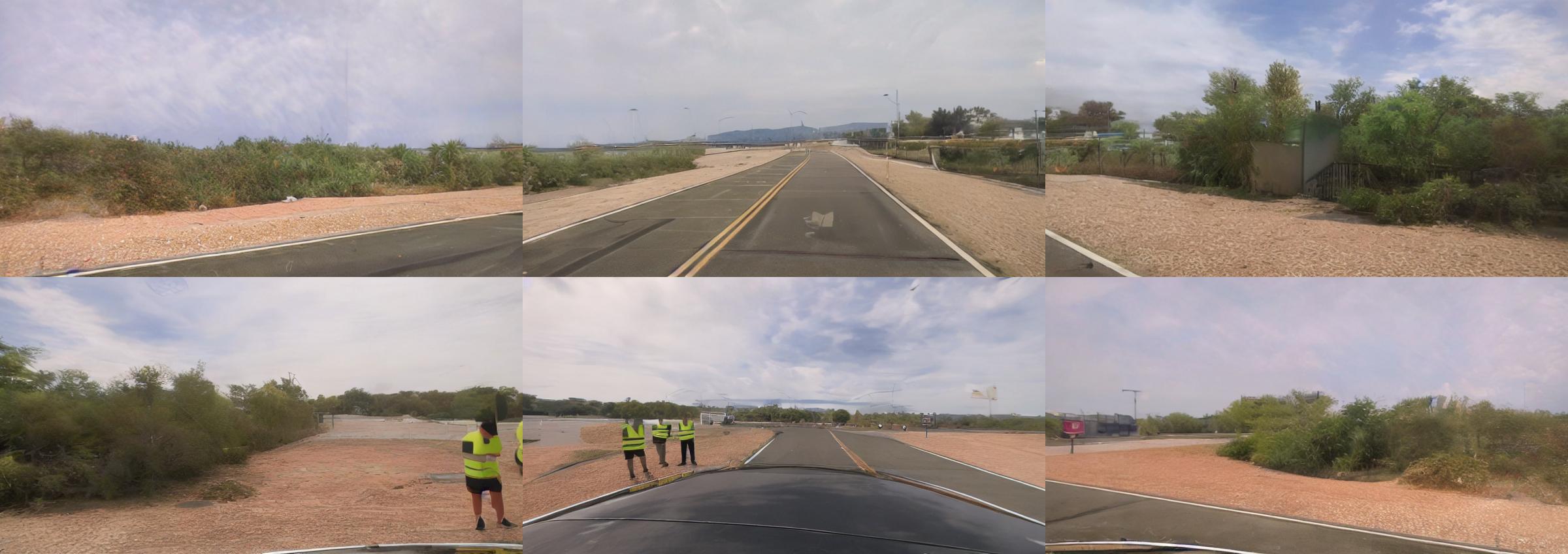}
    \caption{\sysname optimized output, boundary removal.}
  \end{subfigure}\hfill
  \begin{subfigure}{0.48\textwidth}
    \centering
    \includegraphics[width=\linewidth]{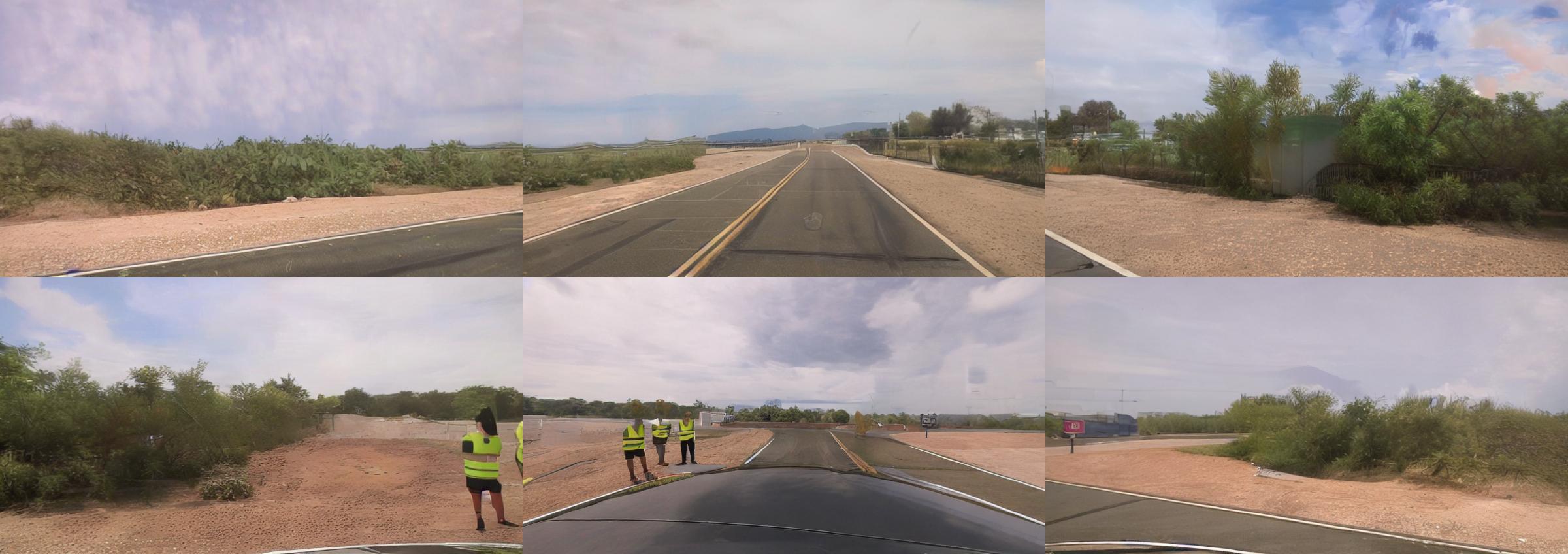}
    \caption{\sysname optimized output, boundary injection.}
  \end{subfigure}
  \begin{subfigure}{0.48\textwidth}
    \centering
    \includegraphics[width=\linewidth]{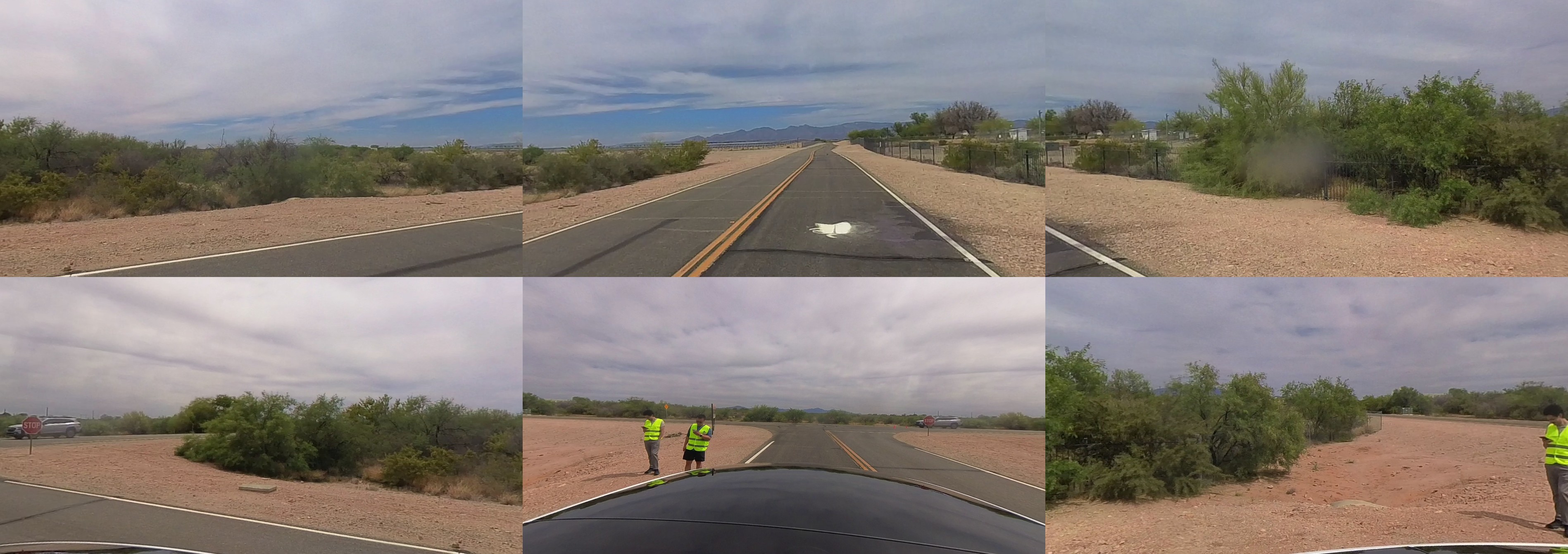}
    \caption{Chalk-reproduced view, boundary removal.}
  \end{subfigure}\hfill
  \begin{subfigure}{0.48\textwidth}
    \centering
    \includegraphics[width=\linewidth]{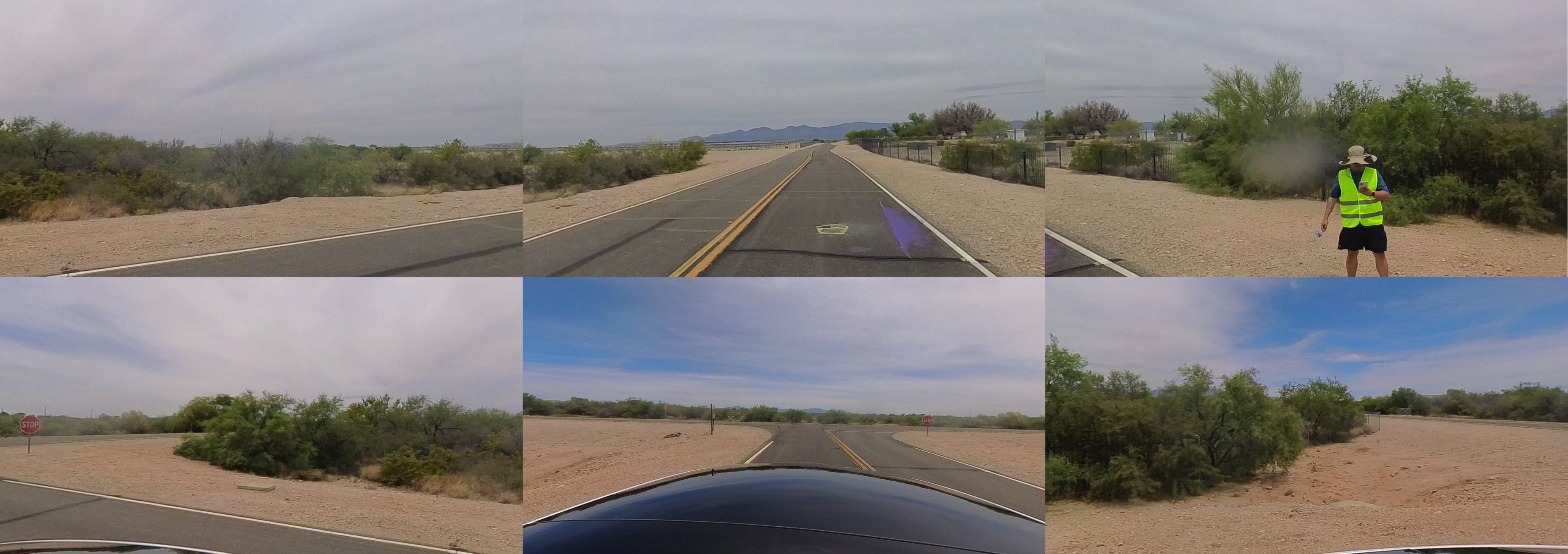}
    \caption{Chalk-reproduced view, boundary injection.}
  \end{subfigure}
  \caption{\textbf{Camera evidence for the physical case study.} The chalk reproduction captures only a coarse outline of the optimized pattern's shape and location, with no attempt to match texture, edge softness, or color.}
  \Description{Two rows of six-camera surround driving views: two adversarial views generated by Mirage (boundary removal and injection) and two physical re-captures of the road after chalk markings were applied.}
  \label{fig:rwf_views_surround}
\end{figure}

\input{tabs/tab_realism_examples}

\section{VLM Realism Prompt and Qualitative Samples}
\label{app:realism_prompt}

Both VLM judges (InternVL3-8B-Instruct and Gemma-4-E4B-it) receive the same expert-persona system prompt and 4-part user prompt. Responses are parsed into a structured schema with fields for scene assessment, realism indicators, YES/NO verdict, and 1--5 confidence score.

\paragraph{System prompt.} \small\emph{``You are an expert in autonomous driving perception and computer vision. Your task is to assess whether a given driving scene image could plausibly have been captured by a real vehicle-mounted camera under real-world conditions. Focus on the physical and semantic realism of the scene content---road markings, lane lines, curbs, signage, lighting, and shadows---not on JPEG compression artifacts or camera sensor noise, which are expected in real dashcam footage.''}

\normalsize
\paragraph{User prompt.} \small\emph{``Look carefully at this image from a front-facing or surround-view camera on a driving vehicle. Answer the following:
\begin{enumerate}
\item SCENE ASSESSMENT: Describe in 1--2 sentences what you observe in the scene (road layout, markings, lighting conditions, any notable features).
\item REALISM INDICATORS: List up to three visual cues that support or undermine the physical realism of the scene. Consider whether road markings and lane geometry follow plausible real-world patterns, whether lighting, shadows, and surface textures are physically consistent, and whether any regions exhibit unnatural patterns, repetitive textures, geometric noise overlays, or high-frequency perturbations inconsistent with natural scene variation.
\item VERDICT: Based on the above, would a human driving in this environment plausibly encounter this exact scene? Answer with exactly one word on its own line: YES or NO.
\item CONFIDENCE: Rate your confidence in the verdict on a scale of 1--5 (1 = very uncertain, 5 = very certain).
\end{enumerate}''}

We provide 8 representative samples spanning all five categories, showing verbatim VLM outputs. Each sample reports both judges' parsed responses.

%% file: tabs/tab_realism_examples.tex
\begin{table*}[!t]
\centering
\footnotesize
\renewcommand{\arraystretch}{1.15}
\setlength{\tabcolsep}{4pt}
\caption{\textbf{VLM realism qualitative samples.} For each image, both judges' parsed responses are shown verbatim. Each VLM cell reports the YES/NO \emph{verdict}, \emph{confidence} on a 1--5 scale, and the free-form \emph{realism indicators} rationale. Across all four image categories, the two judges agree on most samples, with the AdvPatch row showing a rare case of disagreement on a sample where the patch is partially obscured.}
\label{tab:vlm_realism_examples}
\begin{tabular}{@{}>{\centering\arraybackslash}m{0.07\textwidth} >{\centering\arraybackslash}m{0.16\textwidth} m{0.34\textwidth} m{0.34\textwidth}@{}}
\toprule
\multirow{2}{*}{\textbf{Type}} & \multirow{2}{*}{\textbf{Image}} & \multicolumn{2}{c}{\textbf{Verdict, Confidence (1--5), and Realism Indicators}} \\
\cmidrule(lr){3-4}
 & & \textbf{InternVL3} & \textbf{Gemma-4} \\
\midrule
\multirow{2}{*}{\shortstack{Clean\\nuScenes}}
 & \includegraphics[width=\linewidth]{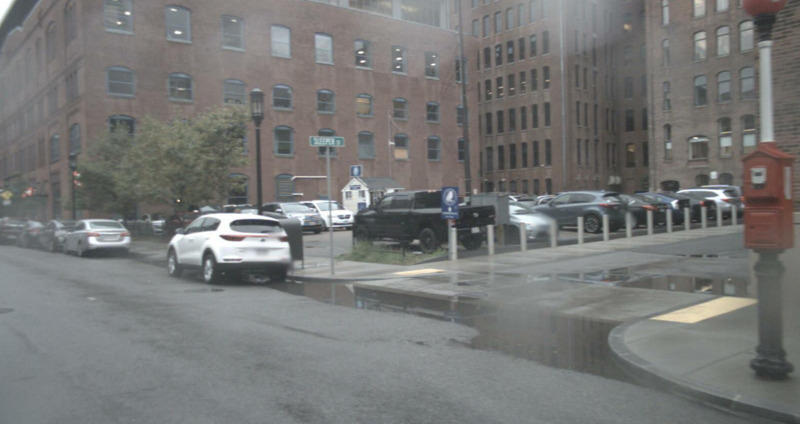}
 & \textbf{YES (5/5).} \emph{``Road markings consistent with typical urban street patterns. Lighting and shadows consistent with an overcast day, wet road surface reflects light naturally. No obvious signs of unnatural patterns or repetitive textures.''}
 & \textbf{YES (5/5).} \emph{``Plausible urban setting; the density of buildings, the presence of street infrastructure, and the wet road surface are highly consistent with a real city environment. Consistent lighting/weather; diffuse low-contrast lighting and wet sheen on asphalt are physically consistent with rainy or overcast weather.''} \\
\cmidrule(l){2-4}
 & \includegraphics[width=\linewidth]{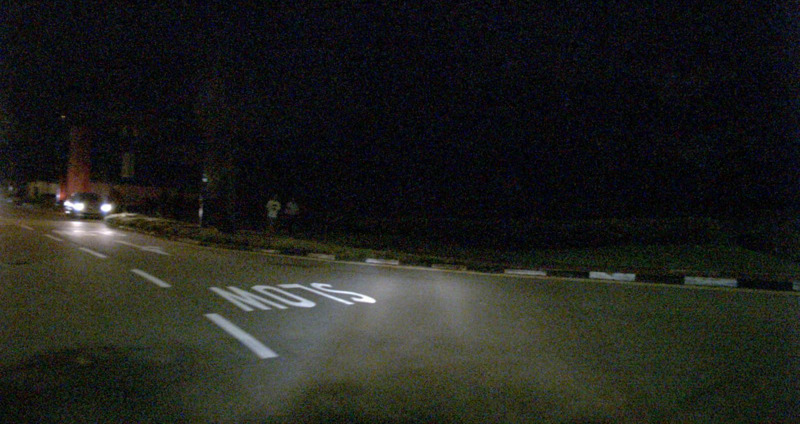}
 & \textbf{NO (4/5).} \emph{``The `MOTS' sign is unusual and may indicate a specific location or context, which could be less common.''}
 & \textbf{NO (4/5).} \emph{``The large, perfectly centered, and brightly illuminated word `SLOW' painted directly on the road surface appears highly stylized and unnatural for standard road signage or markings.''} \\
\midrule
\multirow{2}{*}{PGD}
 & \includegraphics[width=\linewidth]{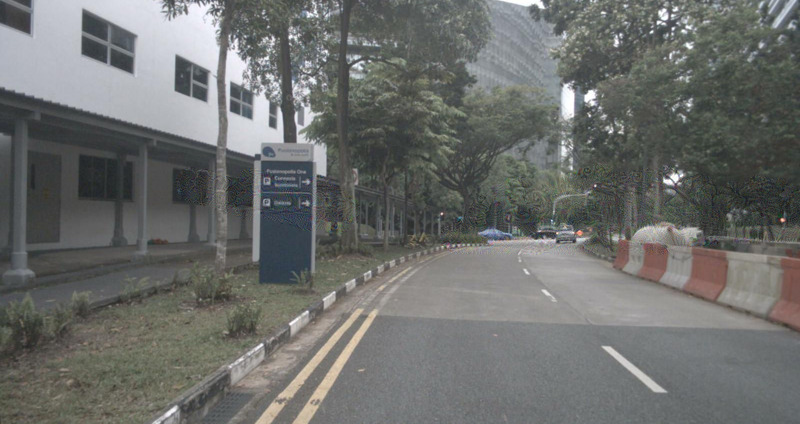}
 & \textbf{YES (5/5).} \emph{``The scene lacks any high-frequency perturbations or unnatural textures that would suggest digital manipulation.''}
 & \textbf{YES (5/5).} \emph{``The road curves naturally, lane markings consistent with standard road painting. Trees and surrounding buildings appear structurally and texturally realistic for an urban environment.''} \\
\cmidrule(l){2-4}
 & \includegraphics[width=\linewidth]{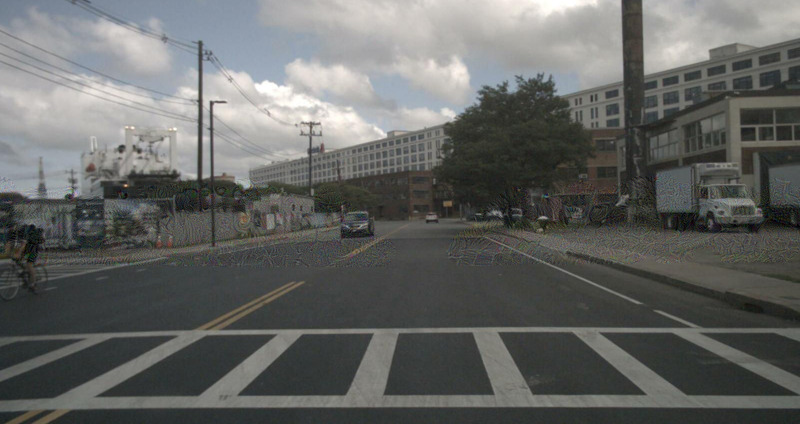}
 & \textbf{NO (4/5).} \emph{``The image exhibits noticeable geometric noise and repetitive textures, particularly in the road markings and background elements, which are inconsistent with natural scene variation.''}
 & \textbf{NO (5/5).} \emph{``There is a noticeable, high-frequency, patterned noise overlay across the entire image, which is inconsistent with natural camera capture and suggests digital manipulation or artifacting.''} \\
\midrule
\multirow{2}{*}{AdvPatch}
 & \includegraphics[width=\linewidth]{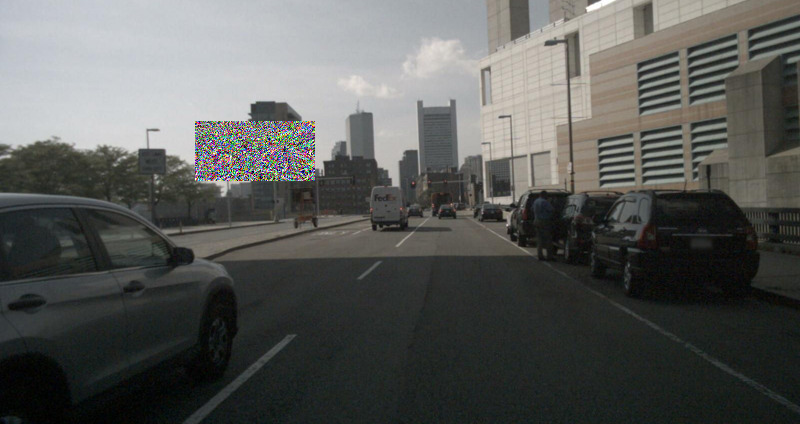}
 & \textbf{NO (4/5).} \emph{``There is a noticeable pixelated or noise artifact in the center of the image. The presence of the pixelated area is an unnatural pattern that undermines the physical realism.''}
 & \textbf{NO (5/5).} \emph{``A significant portion of the image (upper left quadrant) is obscured by a block of random, high-frequency, multi-colored noise, which is a clear indicator of digital corruption or artifacting, not a real-world visual feature.''} \\
\cmidrule(l){2-4}
 & \includegraphics[width=\linewidth]{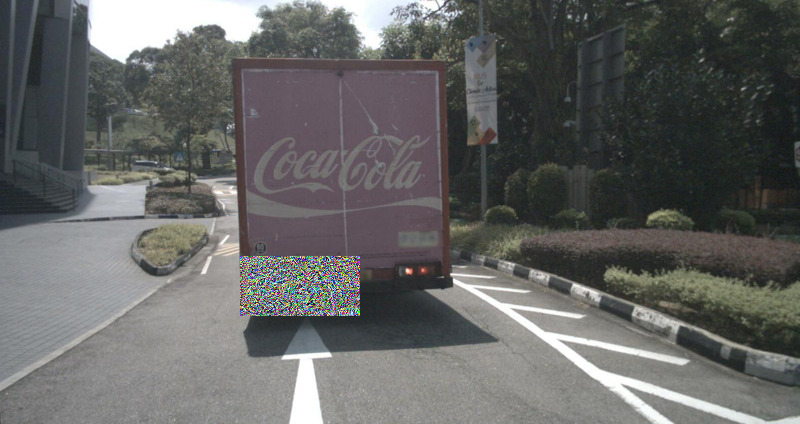}
 & \textbf{YES (4/5).} \emph{``Road markings and lane geometry appear consistent with typical urban or suburban road layouts. Lighting and shadows are consistent with natural daylight conditions, though the image has noticeable JPEG compression artifacts. The truck and surrounding environment have repetitive textures and geometric noise, suggesting digital manipulation or compression effects.''}
 & \textbf{NO (5/5).} \emph{``The road markings and lane geometry appear generally plausible for a suburban or commercial area. The presence of a large, clearly visible Coca-Cola branded truck is a common real-world sight. The image contains significant, patterned noise/overlay artifacts (especially on the left and right edges), which are inconsistent with natural camera capture of a real scene.''} \\
\midrule
\multirow{2}{*}{\sysname}
 & \includegraphics[width=\linewidth]{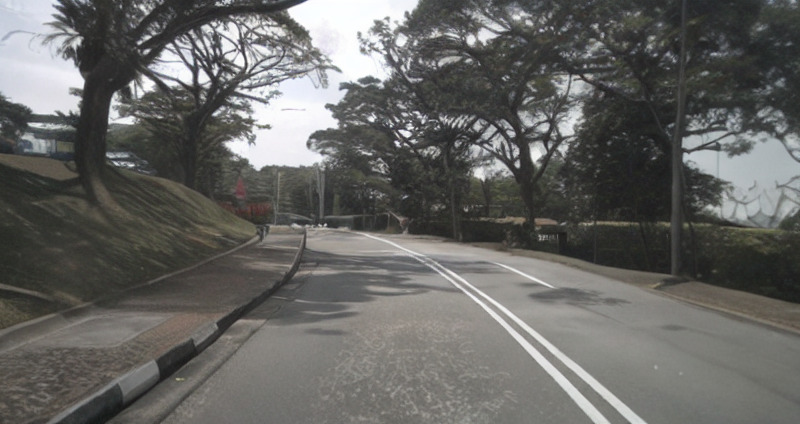}
 & \textbf{YES (5/5).} \emph{``Road markings and lane geometry appear consistent with typical road designs. Lighting and shadows are naturalistic, with no unusual patterns or distortions. Road surface and surrounding vegetation look realistic without any obvious digital artifacts.''}
 & \textbf{YES (5/5).} \emph{``Plausible road geometry; the road appears to follow a natural curve, and the lane markings are consistent with standard road delineation. Consistent lighting/shadows; lighting is diffuse, consistent with an overcast day. Natural textures; textures of asphalt, grass, and tree foliage appear organic and non-synthetic.''} \\
\cmidrule(l){2-4}
 & \includegraphics[width=\linewidth]{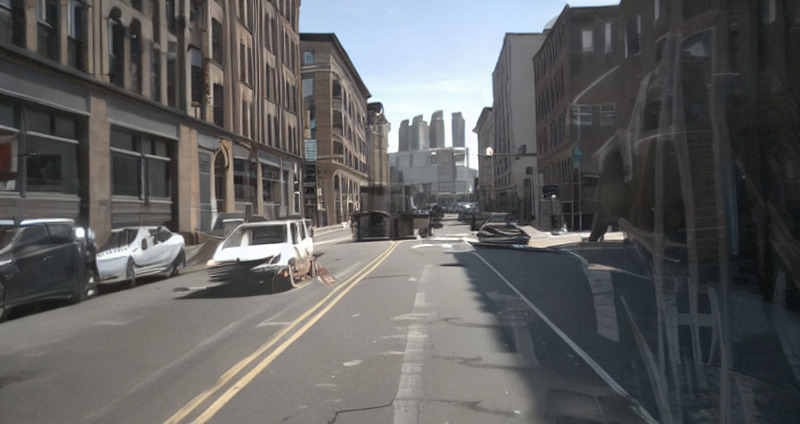}
 & \textbf{NO (4/5).} \emph{``The image has a noticeable distortion on the right side, suggesting a possible overlay or artifact.''}
 & \textbf{NO (5/5).} \emph{``There are significant, large-scale, semi-transparent geometric overlays or artifacts visible across the right side of the image, which are inconsistent with a natural camera capture of a real-world scene.''} \\
\bottomrule
\end{tabular}
\end{table*}